\documentclass{article}

     \usepackage[final]{neurips_2021}

\usepackage{docmute}
\usepackage[utf8]{inputenc} %
\usepackage[T1]{fontenc}    %
\usepackage{amsmath, amssymb}
\usepackage{algorithmicx,algorithm}
\usepackage{algpseudocode}
\usepackage{hyperref}       %
\usepackage{url}            %
\usepackage{booktabs}       %
\usepackage{amsfonts}       %
\usepackage{nicefrac}       %
\usepackage{microtype}      %
\usepackage{graphicx}
\usepackage{color}
\usepackage{multirow}
\usepackage{ntheorem}
\newtheorem*{proof}{Proof}
\usepackage{xfrac}
\usepackage{algorithmicx,algorithm}
\usepackage{algpseudocode}
\usepackage{amsmath}
\usepackage{wrapfig}
\usepackage{mdframed}
\usepackage{accents}

\newtheorem{theorem}{Theorem} %

\author{%
  Cong Geng,\enspace Jia Wang,\enspace Zhiyong Gao \\
  Shanghai Jiao Tong University \\
  \texttt{\{gengcong, jiawang, zhiyong.gao\}@sjtu.edu.cn} \\
  \And
  Jes Frellsen\thanks{equal contribution},\enspace Søren Hauberg\footnotemark[1] \\
  Technical University of Denmark \\
  \texttt{\{jefr, sohau\}@dtu.dk}
}

\renewcommand{\vec}[1]{\mathbf{#1}}
\newcommand{\mat}[1]{\mathbf{#1}}
\newcommand{\x}[0]{\vec{x}}
\newcommand{\z}[0]{\vec{z}}
\newcommand{\J}[0]{\mat{J}}
\newcommand{\dif}[0]{\mathrm{d}}
\newcommand{\T}[0]{^{\intercal}}

\definecolor{linecolor}{RGB}{236,222,212} %
\newmdenv[
  leftmargin = 0pt,
  innerleftmargin = 0.5em,
  innertopmargin = 0pt,
  innerbottommargin = 0pt,
  innerrightmargin = 0pt,
  rightmargin = 0pt,
  linewidth = 2pt,
  topline = false,
  rightline = false,
  bottomline = false,
  linecolor=linecolor,
  innertopmargin=0em,
]{leftbar}

\usepackage[capitalise,noabbrev]{cleveref}

\usepackage{mathtools}
\DeclarePairedDelimiter\ceil{\lceil}{\rceil}            %
\DeclarePairedDelimiter\floor{\lfloor}{\rfloor}         %

\newcommand{\KL}[2]{\mathrm{KL}\left( #1 \| #2 \right)} %
\renewcommand{\L}[0]{L(\theta)}                         %
\newcommand{\expt}[0]{\mathbb{E}}                       %
\newcommand{\lowerOne}[0]{\mathcal{L}(\theta)}

\setcitestyle{authoryear}

\title{Bounds all around: training energy-based models with bidirectional bounds}

\begin{document}

\maketitle

\begin{abstract}
  Energy-based models (EBMs) provide an elegant framework for density estimation, but they are notoriously difficult to train. Recent work has established links to generative adversarial networks, where the EBM is trained through a minimax game with a variational value function. We propose a bidirectional bound on the EBM log-likelihood, such that we maximize a lower bound and minimize an upper bound when solving the minimax game. We link one bound to a gradient penalty that stabilize training, thereby provide grounding for best engineering practice. To evaluate the bounds we develop a new and efficient estimator of the Jacobi-determinant of the EBM generator. We demonstrate that these developments significantly stabilize training and yield high-quality density estimation and sample generation.\looseness=-1
\end{abstract}

\section{Energy-based models}
    Energy-based models (EMBs) are probabilistic models that draw inspiration from physics and have a long history in machine learning \citep{Hopfield2554,hinton1983optimal,10.5555/104279.104290}. An EBM is specified in terms of an energy function $E_{\theta}: \mathcal{X} \rightarrow \mathbb{R}$ that is parameterized by $\theta$ and defines a probability distribution over $\mathcal{X}$ from the Gibbs density:
    \begin{equation}
    	p_{\theta}(\x)=\frac{\exp(-E_{\theta}(\x))}{Z_{\theta}},
    	\qquad
    	Z_{\theta}=\int \exp(-E_{\theta}(\x)) \dif \x,
    	\label{eq:ebm}
    \end{equation}
    where $Z_{\theta}$ is the normalization constant or partition function. In principle, any density can be described this way for a suitable choice of $E_{\theta}$. EBMs are typically learned using maximum likelihood estimation (MLE), where we wish to find a value of $\theta$ that minimized the \emph{negative} data log-likelihood:
    \begin{equation}
    	\L := -\expt_{\x \sim p_{\text{data}}(\x)}\left[\log p_{\theta}(\x)\right] = \expt_{\x \sim p_{\text{data}}}[E_{\theta}(\x)] + \log Z_{\theta},
    \end{equation}
    where $p_{\text{data}}$ is the data generating distribution.

    The fundamental challenge with EBMs for non-standard energy functions is the lack of a closed-form expression for the normalization constant, $Z_{\theta}$, which hinders exact pointwise density evaluation, sampling and learning.  Therefore, these tasks are traditionally performed using approximate methods such as Markov chain Monte Carlo (MCMC, \citealp{metropolis1953equation,neal2011mcmc}). Similar to the Boltzmann learning rule \citep{hinton1983optimal, osogami2017boltzmann}, gradient-based learning for MLE involves evaluating the gradient $\nabla_{\theta}\L = \expt_{\x \sim p_{\text{data}}}[\nabla_{\theta}E_{\theta}(\x)] - \expt_{\x \sim p_{\theta}}[\nabla_{\theta}E_{\theta}(\x)]$, where the expectation in the last term necessitates MCMC approximation.
    Recently, a series of papers \citep{kim2016deep, kumar2019maximum, abbasnejad2020gade, che2020your} have established links between EBMs and \emph{Wasserstein GANs} (WGANs, \citealp{arjovsky2017wasserstein}), such that EBMs can be approximately learned through a minimax game. \citet{xie2017synthesizing,xie2018learning} and \citet{wu2018sparse} also employed the adversarial game of EBM. This allows for significant improvements in EBM learning over MCMC based methods \citep{nomcmc}.
    
    \textbf{In this paper}, we remark that variational bounds on the value function of the minimax game can be problematic as this is both maximized and minimized. We propose to both upper and lower bound the negative log-likelihood, and alternate between their respective minimization and maximization to alleviate this concern. Evaluation of the bounds requires evaluating the entropy of the generator, and we provide a new efficient estimator. We link the upper bound to the use of gradient penalties which are known to stabilize training. Experimentally, we demonstrate that our approach matches or surpasses state-of-the-art on diverse tasks at negligible performance increase.

    \subsection{Background: Variational bounds and minimax games}
        
        Following \cite{nomcmc}, we can lower bound the log normalization constant using a proposal (variational) distribution $p_g(\x)$ and Jensen's inequality:
        \begin{equation}
            \log Z_{\theta} = \log \expt_{\x \sim p_g}\! \left[ \frac{\exp(-E_{\theta}(\x))}{p_g(\x)} \right] \geq \expt_{\x \sim p_g}\! \left[ \log \frac{\exp(-E_{\theta}(\x))}{p_g(\x)} \right]
            = -\expt_{\x \sim p_g}[E_{\theta}(\x)] + H[p_g]
            \label{eq:logZbound}
        \end{equation}
        where $H[p_g] = - \expt_{p_g}[\log p_g]$ is the (differential) entropy of the proposal distribution. This means that we also obtain a lower bound on the negative log likelihood function given by
        \begin{equation}
            \lowerOne := \expt_{\x \sim p_{\text{data}}(\x)}[E_{\theta}(\x)] - \expt_{\x \sim p_g(\x)}[E_{\theta}(\x)] + H[p_g] \leq \L .
	        \label{eq:lower1}
        \end{equation}
        We note that this bound is tight when $p_g = p_{\theta}$, since the bound on the log normalization constant~\eqref{eq:logZbound} can be equivalently expressed as $\log Z_{\theta} \geq \log Z_{\theta} - \KL{p_g}{p_{\theta}}$. To tighten the bound, we seek a proposal distribution $p_g$ that maximises $\lowerOne$, while for MLE we want to find a $\theta$ that minimizes $\lowerOne$. Accordingly, MLE can be formulated as a minimax game:
        \begin{equation}
            \min_{E_{\theta}} \max_{p_g} \left\{ \lowerOne \right\}
            = \min_{E_{\theta}} \max_{p_g} \left\{ \expt_{\x \sim p_{\text{data}}(\x)}[E_{\theta}(\x)] - \expt_{\x \sim p_g(\x)}[E_{\theta}(\x)] + H[p_g] \right\}.
            \label{eq:minimax1}
        \end{equation}
        This gives a tractable and MCMC-free approach to EBM learning, assuming we can evaluate the entropy \citep{nomcmc}.
        
        \textbf{Our key issue} is that both minimizing and maximizing a lower bound is potentially unstable. By minimizing a lower bound, we run the risk of finding an optimum, where the bound is loosest, rather than where it is informative about the optima of the true objective. In particular, a minima of the lower bound may be $-\infty$, which is rather unhelpful. Good results have been reported from using a lower bound in the minimax game \citep{dai2017calibrating, kumar2019maximum, nomcmc}, but the conceptual issue remains.
        
        \textbf{The WGAN loss function} is, as noted by \citet{arjovsky2017wasserstein}, very similar to the lower bound $\lowerOne$,\looseness=-1
        \begin{equation}
            L_{\text{WGAN}}
              = \expt_{\x \sim p_{\text {data }}}[E_{\theta}(\x)]
              - \expt_{\x \sim p_g} \left[ E_{\theta}(\x) \right],
        \end{equation}
        with the missing entropy term being the only difference. Lessons from the WGAN can, thus, be expected to apply to the EBM setting as well.
        For example, the success of \emph{gradient clipping}
        \citep{arjovsky2017wasserstein} and \emph{gradient penalties}
        \citep{gulrajani2017improved} in WGAN training, inspired \citet{nomcmc} to
        heuristically use gradient penalties in variational inference
        to stabilize training. Our work will provide justification to such heuristics.

\section{Approximate minimax games through bidirectional bounds}
    \begin{wrapfigure}[9]{r}{0.27\textwidth}
        \vspace{-7mm}
        \begin{center}
        \includegraphics[width=0.27\textwidth]{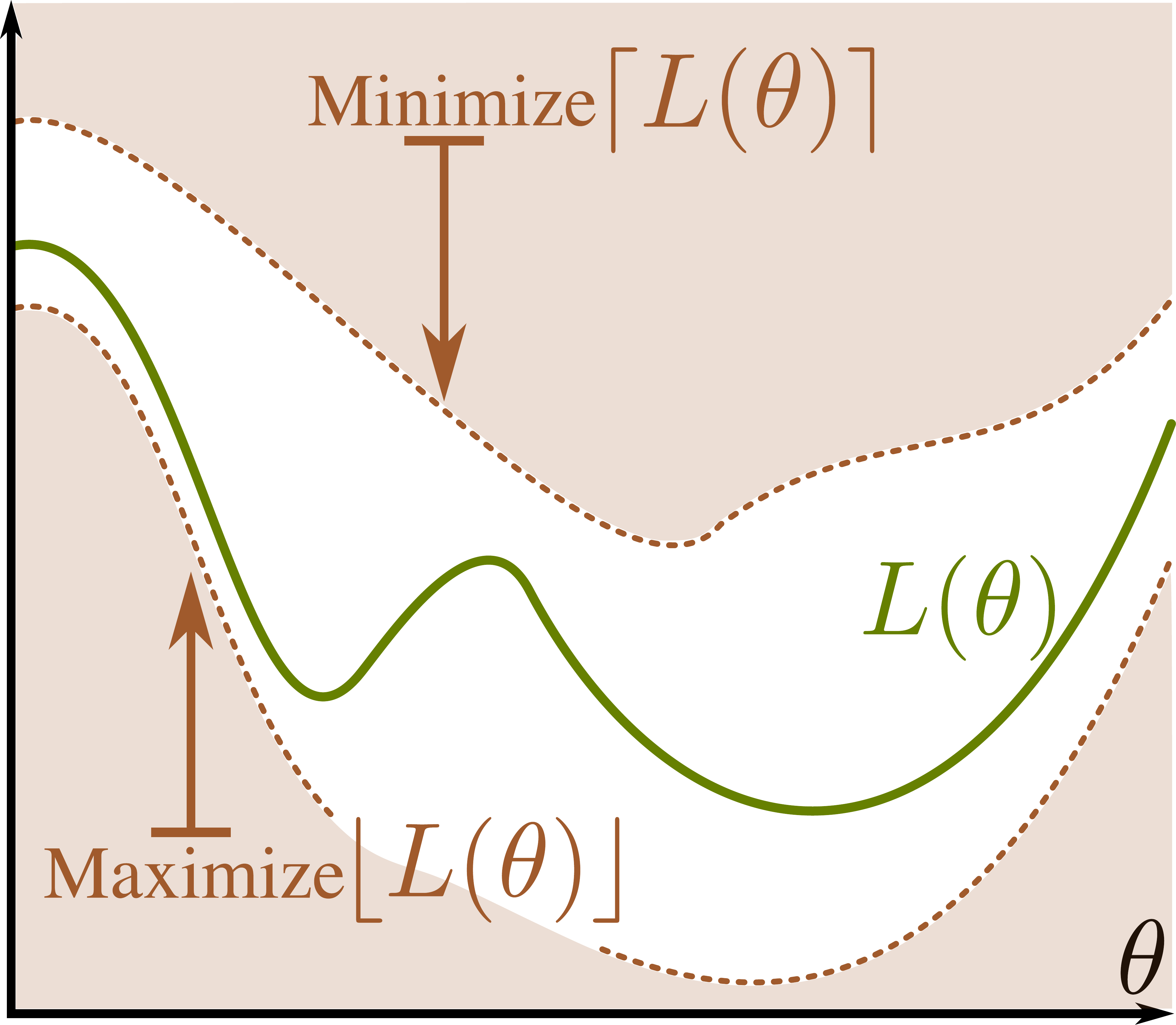}
      \end{center}
      \vspace{-5mm}
      \caption{The bidirectional bounds `sandwich' the negative data log-likelihood.}
      \label{fig:sandwich}
    \end{wrapfigure}
    Instead of solving a minimax game with a lower bounded value function~\eqref{eq:minimax1}, we propose to bound $\L$ from both above and below:
    \begin{equation}
      \floor{\L} \leq \L \leq \ceil{\L}.
    \end{equation}
    With this, we can now follow an optimization strategy of alternating
    \begin{enumerate}
      \item minimize $\ceil{\L}$ with respect to $E_{\theta}$.
      \item maximize $\floor{\L}$ with respect to $p_g$.
    \end{enumerate}
    This avoids the potential pitfalls of minimizing a lower bound.

    \subsection{A lower bound for maximizing $p_g$}
        We already have a lower bound $\lowerOne$ on $\L$ in \cref{eq:lower1}. The first two terms of the bound can readily be evaluated by sampling from $p_{\text{data}}$ and $p_g$, respectively. The entropy of $p_g$ is less straight-forward. Following standard practice in GANs, we define $p_g$ through a base variable $\z \sim \mathcal{N}(\vec{0}, \mat{I})$, which is passed through a `generator' network $G: \mathbb{R}^d \rightarrow \mathbb{R}^D$, i.e.\@ $\x = G(\z) \sim p_g$.
        This constructs a density over a $d$-dimensional manifold\footnote{Note that Eq.~\ref{eq:log_p_g} is only the density \emph{on} the spanned manifold, and that the density is zero \emph{off} the manifold.}
        in $\mathbb{R}^D$, which by standard change-of-variables is
        \begin{equation}
            \log p_g (G(\z)) = \log p_0 (\z) - \frac{1}{2}\log\det(\J_{\z}\T \J_{\z}),
            \label{eq:log_p_g}
        \end{equation}
        where $\J_{\z} \in \mathbb{R}^{D \times d}$ is the Jacobian of $G$ at $\z$.
        This assume that the Jacobian exist, i.e.\@ that the generator $G$ has at least one derivative. 
        The entropy of $p_g$ is then
        \begin{equation}
            H[p_g]
              = -\expt_{\x \sim p_g}\left[ \log p_g(\x) \right]
               = H[p_0] + \expt_{\z \sim p_0}\left[ \frac{1}{2}\log\det(\J_{\z}\T \J_{\z}) \right].
        \end{equation}
        The entropy of $p_0$ trivially evaluates to $H[p_0] = \sfrac{d}{2}(1 + \log(2\pi))$. To avoid an expensive evaluation of the log-determinant-term, we note that it is easily bounded from below:
        \begin{equation}
            \frac{1}{2} \log\det(\J_{\z}\T \J_{\z})
              = \frac{1}{2} \sum_{i=1}^d \log s_i^2
               \geq d \log s_1,
        \end{equation}
        where $s_d \geq \ldots \geq s_1$ are the singular values of the Jacobian $\J_{\z}$. Note that in general, this bound is not tight.
        With this, we have our final lower bound:
        \begin{equation}
            \L \geq \floor{\L}
                = \expt_{\x \sim p_{\text{data}}(\x)}[E_{\theta}(\x)]
	            - \expt_{\x \sim p_g(\x)}[E_{\theta}(\x)]
	            + H[p_0]
	            + \expt_{\z \sim p_0}\left[ d \log s_1(\z) \right].
	        \label{singular}
       \end{equation}
        
    \subsection{An upper bound for minimizing $E_\theta$}
        There are many ways to arrive at an upper bound for the log-likelihood. Due to the strong ties between the lower bound~\eqref{eq:lower1} and the WGAN objective, we take inspiration from how WGANs are optimized. The original WGAN paper \citep{arjovsky2017wasserstein} noted optimization instabilities unless \emph{gradient clipping} was applied. \citet{gulrajani2017improved} noted that this clipping could be avoided by adding a \emph{gradient penalty}, such that the loss become
        \begin{align}
            L_{\text{WGAN-GP}}
              &= L_{\text{WGAN}}
	           + \lambda \expt_{\hat{\x} \sim g(\x)} \left[(\|\nabla_{\hat{\x}}E_{\theta}(\hat{\x})\|_2 - 1)^2\right],
        \end{align}
        where $\hat{x}$ is a uniform interpolation between real data and generated samples. This is generally acknowledged as a good way to train a WGAN \citep{gulrajani2017improved}. We observe that this loss is similar to our lower bound, but the gradient penalty adds a positive term, such that one could speculate that we might have an upper bound.
        With this intuition in hand, we prove the following statement in the supplementary material.\smallskip%
        \begin{leftbar}
        \begin{theorem}\label{thm:upper} 
            Suppose $f: \mathcal{X} \rightarrow \mathbb{R}$ is L-Lipschitz continuous, $g(x)$ is a probability density function with finite support, then there exists constants $M, m \geq 0$ and $p \geq 1$ such that:
            \begin{equation}
                \log \expt_{\x \sim g(\x)} \left[f(\x)\right]-\expt_{\x \sim g(\x)} \left[\log f(\x)\right] \leq M (\expt_{\x \sim g(\x)} \left[\vert\nabla_{\x}\log f(\x)\vert^p\right]+m)^{\sfrac{1}{p}}.
                \label{eq:thm_upper}
            \end{equation}
        \end{theorem}
        \end{leftbar}
        Note that if $\expt_{\x \sim g(\x)} \left[\vert\nabla_{\x}\log f(\x)\vert^p\right]+m \geq 1$, the bound in \cref{eq:thm_upper} can be simplified by dropping the power $\sfrac{1}{p}$,
        \begin{equation}
            \log \left[\expt_{\x \sim g(\x)} \left[f(\x)\right]\right] - \expt_{\x \sim g(\x)} \left[\log f(\x)\right]
            \leq
            M\expt_{\x \sim g(\x)} \left[\vert\nabla_{\x}\log f(\x)\vert^p\right] + Mm.
            \label{eq:thm_upper_simple}
        \end{equation}
        For our model, we set $f(\x) = \frac{\exp(-E_{\theta}(\x))}{p_g(\x)}$ and $g(\x)=p_g(\x)$. Empirically, we observe the bound simplification in \cref{eq:thm_upper_simple} holds through-out most of the training, and use the following upper bound:
        \begin{equation}
            \ceil{\L}
              = \lowerOne
              + M\expt_{\x \sim p_g(\x)} \left[\vert \nabla_{\x} E_{\theta}(\x) + \nabla_{\x}\log p_g(\x)\vert^p\right]
              + Mm.
            \label{eq:upper}
        \end{equation}
        For a given choice of $f(\x)$, we can view $m$ as a constant and the term $Mm$ can be ignored during optimization. This upper bound can directly be interpreted as the lower bound \eqref{eq:lower1} plus a gradient penalty, albeit one of a different form than the traditional WGAN penalty, which is derived purely from a regularization perspective. Our upper bound can, thus, be seen as a justification of the regularization from a maximum likelihood perspective.
        
        \paragraph{Bound tightness} When $p_\theta = p_g$, we have that $\left|\nabla_{\x}E_{\theta}(\x)+\nabla_{\x} \log p_{g}(\x)\right|^{p} = 0$. In \cref{thm:upper}, $m$ is a constant related to the Lipschitz constant of $\log f(\x)$ satisfying $\left|\nabla_{\x} \log f(\x)\right|^{p} \leq\left|\nabla_{\vec{y}} \log f(\vec{y})\right|^{p}+m$ for all $\x,\vec{y}$ (see proof for details). When $p_\theta = p_g$ we also have $\left|\nabla_{\x} \log f(\x)\right|^{p} = 0$, such that $m=0$. Our upper bound is then $\ceil{\L} = \lowerOne=\L$, and hence it is tight.

\section{Numerical evaluation of the bounds}
    \paragraph{Evaluating the lower bound}
    To evaluate the lower bound in \cref{singular}, we need the smallest singular value of the Jacobian $\J = \partial_{\z} G(\z)$. Recall that this singular value satisfy $s_1 = \|\J \vec{v}_{\text{min}} \|_2 = \min_{\vec{v} \neq \vec{0}} \frac{\| \J\vec{v}\|_2}{\|\vec{v}\|_2}$. We can then evaluate the singular value by finding $\vec{v}_{\text{min}}$ with an iterative optimization algorithm, where we opt to use the celebrated single-vector \emph{LOBPCG algorithm} \citep{knyazev1998preconditioned}. This method performs an iterative minimization of the generalized Rayleigh quotient,
    \begin{align}
        \rho(\vec{v}) := \frac{\vec{v}\T \J\T \J \vec{v}}{\vec{v}\T \vec{v}},
        \label{eq:rayleigh}
    \end{align}
    which converges to $\vec{v}_{\text{min}}$. The gradient of $\rho(\vec{v})$ is proportional to $r =\J\T \J \vec{v} - \rho(\vec{v}) \vec{v}$. To avoid computing the Jacobian $\J$, we use Jacobian-vector products, which can be efficiently evaluated using automatic differentiation. To compute $\J\T \J \vec{v}$, we use the following trick (in \texttt{pytorch}-notation):
    \begin{align}
        \J\T \J\vec{v}
          = \left((\J\vec{v})\T \J\right)\T
          = \nabla_{\z} \left( (\J\vec{v})\T\!\!\!\texttt{.detach()} \cdot G(\z) \right)\T.
        \label{pro}
    \end{align}
    The optimal learning rate for this iterative scheme can be found by maximizing the Rayleigh quotient~\eqref{eq:rayleigh}. Finally, we follow the suggestions of \citet{knyazev2001toward} to improve numerical stability and accelerate convergence, which we omit here for brevity. 

    \paragraph{Evaluating the upper bound}
    There are two challenges when evaluating \cref{eq:upper}. The first is to compute $\nabla_\x\log p_g(\x)$, where we empirically found that existing methods \citep{shi2018spectral,li2017gradient} were too inefficient for our needs. To evaluate the term $\expt_{\x \sim p_g(\x)} \left[\vert \nabla_{\x} E_{\theta}(\x) + \nabla_\x\log p_g(\x)\vert^p\right]$, we further loosen the bound
    \begin{equation}
    \begin{aligned}
        \vert \nabla_{G(\z)} E_{\theta} (G(\z)) + \nabla_{G(\z)}\log p_g(G(\z))\vert^p
          &\leq \frac{\vert \nabla_{G(\z)}E_{\theta} (G(\z))\J_{\z}
           + \nabla_{G(\z)}\log p_g(G(\z))\J_{\z}\vert^p}{s_1^p}\\
          &\leq \frac{\vert \nabla_{G(\z)} E_{\theta} (G(\z))\J_{\z} + \nabla_\z\log p_g(G(\z))\vert^p}{s_1^p}.
          \label{tranfer}
    \end{aligned}
    \end{equation}
    where $s_1$ is the smallest singular value of $\J_{\z}$. Detailed derivations are in the supplementary material. If we choose $p=2$, then we can use \citeauthor{hutchinson1989stochastic}'s estimator \citeyearpar{hutchinson1989stochastic}:
    \begin{equation}
        \vert \nabla_{\x} E_{\theta} (\x)\J_{\z} + \nabla_\z\log p_g(G(\z)) \vert^2
          = \expt_{\vec{v}}\left[ \left(\nabla_{\x} E_{\theta} (\x)\J_{\z}\vec{v}
          + \nabla_\z\log p_g(G(\z))\vec{v}\right)^2 \right],
        \label{eq:gra_esti}
    \end{equation}
    where $\vec{v} \sim \mathcal{N}\left(\vec{0}, \mat{I}_{d}\right)$.
    This is easily evaluated using automatic differentiation.

    The second challenge is to evaluate $\log p_g(\x)$ which needs the Jacobian of the generator $G(\z)$ as dictated by \cref{eq:log_p_g}.
    Here, we opt to use our entropy estimator as described above. We could alternatively use Hutchinson's estimator as proposed by \citet{kumar2020regularized}. Experimentally we do not observe much difference between these two estimators.

\section{Related work}

In machine learning, there has been a long-standing interest in EBMs dating back to Hopfield networks \citep{Hopfield2554}, Boltzmann machines \citep{hinton1983optimal,hinton1985learning} and restricted Boltzmann machines \citep{10.5555/104279.104290,hinton2002training}, see e.g.\@ reviews in the works by \citet{lecun2006tutorial} and \citet{scellier2020deep}. Learning and evaluation of these models are difficult since the normalization constant cannot be efficiently evaluated. MLE-based learning, such as the Boltzmann learning rule, relies on expensive MCMC sampling to estimate the gradient, and more advanced MCMC methods are used to reliably estimate the normalization constant (see e.g.\@ \citealp{salakhutdinov:icml08a,NIPS2013_fb60d411,pmlr-v38-burda15,pmlr-v51-frellsen16}). For images, MCMC-based learning has been used to learn non-deep EBMs of both textures \citep{zhu1998filters,zhu1998grade} and natural images \citep{xie2015learning,xie2016inducing}. Learning algorithms that avoid the costly MCMC approximation have been heavily investigated. For instance, \citet{hinton2002training} proposed k-step Contrastive Divergence (CD-k) to approximate the negative phase log-likelihood gradient, and \citet{hyvarinen2005estimation} proposed an alternative method to train non-normalized graphical models using score matching. Deep versions of EBMs have subsequently been proposed, such as deep belief networks \citep{hinton2006fast} and deep Boltzmann machines \citep{pmlr-v5-salakhutdinov09a}.\looseness=-1

In recent years, there have been renewed interest in deep generative EBMs, particularly for image generation. \citet{nomcmc} gave an excellent overview of the current developments and distinct drawbacks of three classes of learning methods, which we summarize here: (1)~MLE methods with MCMC sampling are slow and unstable, and while (2)~score matching based methods are comparable faster, they are also unstable and do not work with discontinuous nonlinearities. (3)~Noise-contrastive estimation \citep{gutmann2010noise} do not have these drawback, but it does not scale well with the data dimensionality.

Our proposed method belongs to a fourth class of algorithms that sidestep the costly MCMC sampling by using a simultaneously learned generator or variational distribution. 
\citet{kim2016deep} proposed using a generator function and an adversarial-like training strategy similar to ours. They update the generator using the same lower bound as us, but their entropy approximation is quite different. Furthermore, their method does not have a gradient penalty like our upper bound when optimizing the energy function. Consequently, the energy function needs to be explicitly designed to prevent it from growing to infinity, limiting its potential.

The method proposed by \citet{zhai2016generative} plays a min-max game to jointly optimize the energy function and generator using the same lower bound as us. However, their method relies on a specific designed bounded multi-modal energy function, which limits its potential. Furthermore, their approximation of the generator entropy comes with no theoretical guarantees, and their regularisation of the energy function does not constitute an upper bound.

\citet{dai2017calibrating} proposed an adversarial learning framework for jointly learning the energy function and a generator. They considered two approaches to maximising the generator entropy: One, which maximizes the entropy by minimizing the conditional entropy using a variational upper bound, and another, which makes isotropic Gaussian assumptions for the data, which is not suitable for high-dimensional data.
\citet{kumar2019maximum} and \citet{abbasnejad2019generative} also consider adversarial learning but different approaches to estimating the entropy of the generator.
\citet{kumar2019maximum} estimated the entropy through its connection to mutual information, but they need an additional network to measure the entropy term. 
\citet{abbasnejad2019generative} maximized the entropy by approximating the generator function’s Jacobian log-determinant. However, their method is impractical in high dimensions as the Jacobian is computationally expensive.
\citet{han2019divergence} use an adversarial learning strategy in their divergence triangle loss, but their training mechanism is radically different from ours, and they rely on an extra encoder for learning the generator.

\citet{xie2018cooperative, xie2018cooperative2, xie2021cooperative, xie2021learning} proposed \emph{cooperative learning} of the energy function and a generator. However, the cooperative learning approach relies on MCMC or Langevin dynamics to draw samples from the EBM, which is expensive and difficult to tune.
Like our work, VERA by \citet{nomcmc} avoids the use of MCMC. VERA plays a min-max game and uses variational inference to approximate the gradient of the entropy term, which is different from ours. Furthermore, VERA uses a gradient penalty as a regularizer for the energy function, which is a heuristic unlike our upper bound, and their method is memory-consuming, and the hyperparameters are difficult to adjust.

\section{Experiments}
    \label{exp}

    To demonstrate the efficiency of our energy-based model with bidirectional bounds (EBM-BB) we compare against a range of methods that represent state-of-the-art. As representatives of the GAN literature, we consider \emph{deep convolutional GANs (DCGANs)} by \citet{radford2015unsupervised}, \emph{spectrally normalized GANS (SNGANs)} by \citet{miyato2018spectral} and the \emph{WGAN with zero-centered gradient penalty (WGAN-0GP)} by \citet{thanh2019improving}. As representatives of EBMs, we consider the \emph{maximum entropy generators (MEG)} by \citet{kumar2019maximum} and the \emph{variational entropy regularized approximate maximum likelihood (VERA)} estimator \citep{nomcmc}. 
    As representatives of CoopNets, we consider two similar methods~\citep{xie2018cooperative,xie2021cooperative} and \emph{EBM-VAE}  \citep{xie2021learning}. We also consider \emph{NCSN} by \citet{song2019generative} and \emph{DDPM} by \citet{ho2020denoising} as representatives of a score matching model and a diffusion model respectively.
    To investigate the influence of our derived upper bound, we introduce another baseline in which we replace our upper bound with a zero-centered gradient penalty. We denote the resulting model \emph{EBM-0GP}.

    \subsection{Training details}
        On toy data and MNIST, all models are realized with multi-layer perceptrons (MLPs), while for natural images, we use the convolutional architecture from \texttt{StudioGAN} \citep{kang2020ContraGAN}. Details of the network architecture are given in the supplementary material.
        All experiments are conducted on a single 12GB NVIDIA Titan GPU using a \texttt{pytorch} \citep{paszke2017automatic} implementation. 
        To improve training, we use a positive margin for our energy function to balance the bounds. Specifically, we let
        \begin{equation}
            \ceil{\L}
              = \floor{\L}
              + \left[ M\expt_{\x \sim p_g(\x)} \left[\vert \nabla_{\x} E_{\theta} (\x)
                     + \nabla_{\x}\log p_g(\x)\vert^p\right] - \zeta \right]_+,
        \end{equation}
        where $[\cdot]_+ = \max(0, \cdot)$ is the usual hinge. In all experiments, $\zeta = 1$ as this allows us to apply the simplified bound in \cref{eq:upper}.

    \subsection{Toy data}
         \begin{figure}[tb]
            \footnotesize
            \centering
            \renewcommand{\tabcolsep}{1pt} \renewcommand{\arraystretch}{0.1} \begin{tabular}{cccccc}
            	\includegraphics[width=0.16\linewidth]{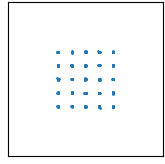} &
            	\includegraphics[width=0.16\linewidth]{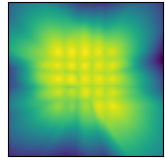} &
            	\includegraphics[width=0.16\linewidth]{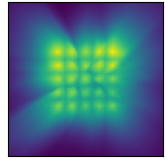} &
            	\includegraphics[width=0.16\linewidth]{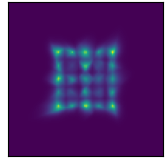}&
            	\includegraphics[width=0.16\linewidth]{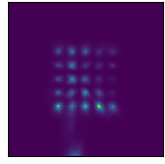} &
            	\includegraphics[width=0.16\linewidth]{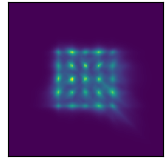} 
            	\\
            	\includegraphics[width=0.16\linewidth]{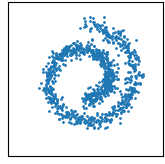} &
            	\includegraphics[width=0.16\linewidth]{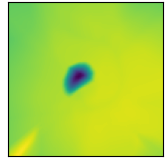} &
            	\includegraphics[width=0.16\linewidth]{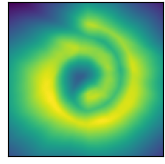} &
            	\includegraphics[width=0.16\linewidth]{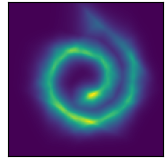}&
            	\includegraphics[width=0.16\linewidth]{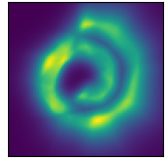} &
            	\includegraphics[width=0.16\linewidth]{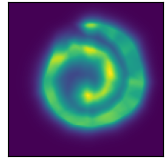} \\
            	(a) Data & (b) WGAN-0GP & (c) MEG & (d) VERA &(e) EBM-0GP~(ours) & (f) EBM-BB~(ours) 
            \end{tabular}
            \caption{Density estimation on the 25-Gaussians and swiss-roll datasets.
              }
            \label{density} 
        \end{figure}
        
        \begin{wraptable}[12]{r}{0.5\linewidth}
            \vspace{-6.5mm}
            \caption{Number of captured modes and KL divergence between the real and generated distributions.}
            \label{mode_counting}
            \vspace{2mm}
            \begin{tabular}{cccccc}
                \toprule
                \multicolumn{2}{c}{Model} & \multicolumn{2}{c}{Modes$\uparrow$} & \multicolumn{2}{c}{KL$\downarrow$} \\ \midrule
                \multicolumn{2}{c}{DCGAN} & \multicolumn{2}{c}{392 $\pm$ 7.4} & \multicolumn{2}{c}{8.012 $\pm$ 0.056} \\
                \multicolumn{2}{c}{SNGAN} & \multicolumn{2}{c}{441 $\pm$ 39.0} & \multicolumn{2}{c}{2.755 $\pm$ 0.033 } \\
                \multicolumn{2}{c}{WGAN-0GP} & \multicolumn{2}{c}{\textbf{1000 $\pm$ 0.0}} & \multicolumn{2}{c}{0.048 $\pm$ 0.003} \\
                \multicolumn{2}{c}{MEG} & \multicolumn{2}{c}{\textbf{1000 $\pm$ 0.0}} & \multicolumn{2}{c}{0.042 $\pm$ 0.004} \\
                \multicolumn{2}{c}{VERA} & \multicolumn{2}{c}{989 $\pm$ 9.0} & \multicolumn{2}{c}{0.152 $\pm$ 0.037} \\
                \multicolumn{2}{c}{EBM-0GP(ours)} & \multicolumn{2}{c}{\textbf{1000 $\pm$ 0.0}} & \multicolumn{2}{c}{\textbf{0.039 $\pm$ 0.003}}\\
                \multicolumn{2}{c}{EBM-BB~(ours)} & \multicolumn{2}{c}{\textbf{1000 $\pm$ 0.0}} & \multicolumn{2}{c}{0.045 $\pm$ 0.003} \\ \bottomrule
            \end{tabular}
        \end{wraptable}
        
        \Cref{density} shows estimated densities on the 25-Gaussians and swiss-roll datasets using both our methods, WGAN-0GP and two baselines.
        We observe that WGAN's discriminator is not suitable as a density estimator. This is unsurprising as WGAN is not supposed to provide a density estimate. MEG and EBM-0GP can have inaccurate density information in some edge and peripheral areas. This may be due to insufficient or excessive maximization of entropy since the zero-centered gradient penalty is not a principled objective for maximum likelihood. VERA and our EBM-BB can learn a sharp distribution, but our method is more stable in some inner regions of the 25-Gaussians and edge regions of the swiss-roll.\looseness=-1

    \subsection{MNIST}
        \paragraph{Mode counting}
        Mode collapse is a frequently occurring phenomenon in which the generator function maps all latent input to a small number of points in the observation space. In particular, GANs are plagued by this problem. Since the generator is trained to maximize the entropy of the generated distribution, several EBM-based models have been shown to capture all the modes of the data distribution. To empirically verify that our model also captures a variety of modes in the data distribution, we follow the test procedure of \citet{kumar2019maximum}. We train our generative model on the StackedMNIST dataset, a synthetic dataset created by stacking MNIST on different channels. The true total number of modes is $1{,}000$, and they are counted using a pretrained MNIST classifier. The KL divergence is calculated empirically between the generated mode distribution and the data distribution. The results appear in \cref{mode_counting}. As expected, GAN-based methods suffer from mode collapse except for WGAN-0GP. All the EBM-based methods capture all or nearly all the modes of the data distribution. Our model EBM-0GP captures all modes and report the smallest KL measures. It worth mentioned that WGAN-0GP obtains comparable results, which was also observed by \citet{nomcmc}. 

        \paragraph{Entropy Estimation}
        \begin{wrapfigure}[11]{r}{0.3\linewidth}
            \vspace{-4mm}
        	\includegraphics[width=\linewidth]{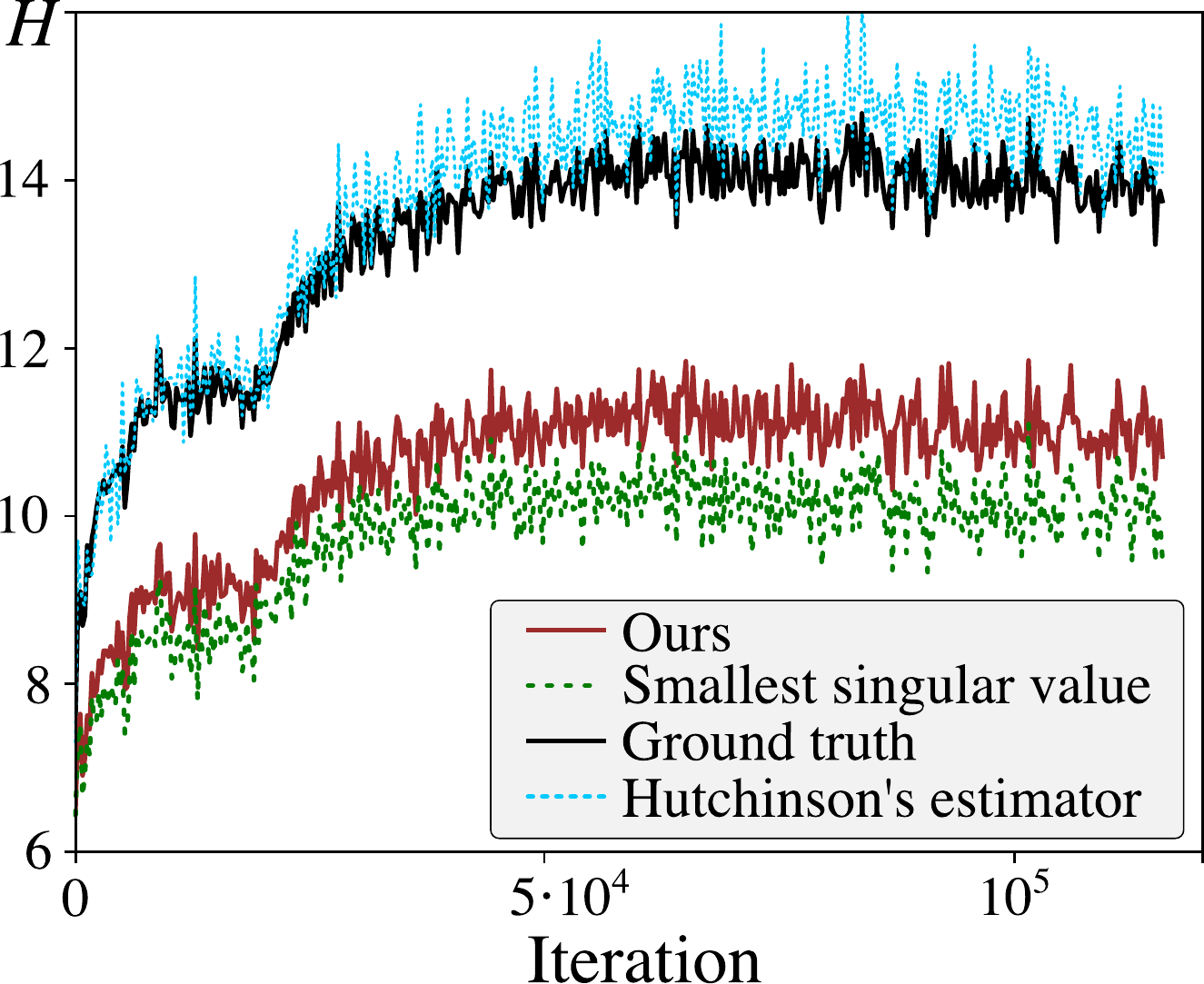}
        	\vspace{-6mm}
        	\caption{Entropy estimators.}
        	\label{entropy estimation}
        \end{wrapfigure}
        In \cref{entropy estimation}, we explore the quality of our entropy estimator measured on MNIST~\citep{lecun1998mnist}. We use networks with fully connected layers as the Jacobian is then easily derived in closed-form, giving us a ground truth for the entropy. We compare our estimator (red) with the ground truth (black) and Hutchinson's estimator (blue), as was proposed by \cite{kumar2020regularized}. Finally, as we only run our iterative estimator for a few steps, we also compare with an estimator that uses the smallest singular value computed with high precision (green).
        We observe that Hutchinson's estimator is reasonable close to the ground truth but provides an upper bound, making it inapplicable for our lower bound. We further observe a noticeable gap between our estimator and the high precision singular value estimator. Finally, we see that all estimators follow roughly the same trend, which suggests that our estimator provides a suitable target for optimization.
   
    \subsection{Natural Images}
        \paragraph{Image generation} The data studied thus far are simple and perhaps do not challenge our upper bound and entropy estimator. We, therefore, train our model on the standard benchmark $32 \times 32$ CIFAR-10~\citep{krizhevsky2009learning} dataset and the $64 \times 64$ cropped ANIMEFACE\footnote{https://www.kaggle.com/splcher/animefacedataset} dataset, which both represent a significant increase in complexity. Similar to recent work on GANs \citep{miyato2018spectral}, we report Inception Score (IS), Fréchet Inception Distance (FID) scores and two $F_\beta$ scores \citep{precision_recall_distributions}. We compare with competitive GANs, EBM baselines, CoopNets, NCSN and DDPM (\cref{image generation}). All GANs and EBMs are reproduced using the same network architecture (see supplements for details) except DCGAN and IGEBM, where we respectively borrow the results from StudioGAN \citep{kang2020ContraGAN} and the original implementation \citep{du2019implicit}. We use the DCGAN~\citep{radford2015unsupervised} network architecture for CIFAR-10 and a Resnet architecture~\citep{kang2020ContraGAN} for ANEMIFACE. Parameters are chosen as in the original papers. For VERA~\citep{nomcmc}, we were unable to reproduce the reported performance, so we choose hyper-parameters according to an extensive grid search. Like the original paper, we choose the entropy weight to be $0.0001$. For CoopNets and NCSN, we report the results from the original papers. For DDPM, we used a public available implementation.\footnote{https://github.com/rosinality/denoising-diffusion-pytorch}
        
\begin{figure*}[tpb]
        	\footnotesize
        	\centering
        	\renewcommand{\tabcolsep}{1pt} \renewcommand{\arraystretch}{0.1} \begin{tabular}{cccc}
        	\vspace{3pt}
        		\includegraphics[width=0.25\linewidth]{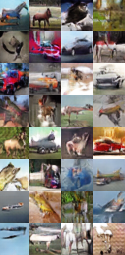} &
        		\includegraphics[width=0.25\linewidth]{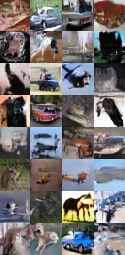} &
        	\includegraphics[width=0.25\linewidth]{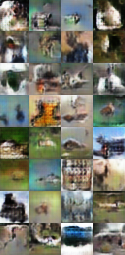} &
        		\includegraphics[width=0.25\linewidth]{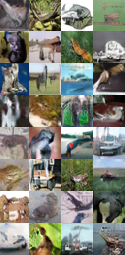} \\ 
        		\vspace{3pt}
        		  (a) WGAN-0GP & (b) MEG & (c) VERA & (d) EBM-BB~(ours) \\
        		 	\vspace{3pt}
        	\end{tabular}
        	\caption{
        	Generated samples on CIFAR-10 with our method and various methods.
        	}
        	\label{cifar10}%
        \end{figure*}

\begin{figure*}[tb]
        	\footnotesize
        	\centering
        	\renewcommand{\tabcolsep}{1pt} \renewcommand{\arraystretch}{0.1} \begin{tabular}{cccc}
        	\vspace{3pt}
        		\includegraphics[width=0.25\linewidth]{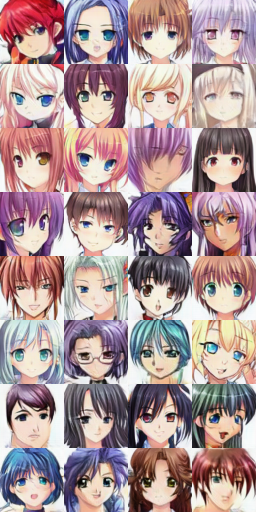} &
        		\includegraphics[width=0.25\linewidth]{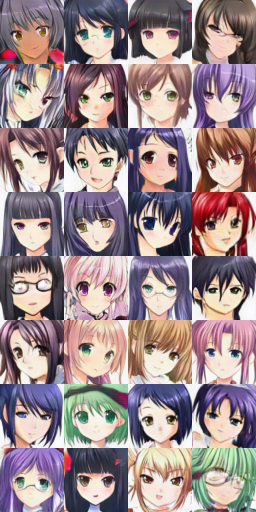} &
        		\includegraphics[width=0.25\linewidth]{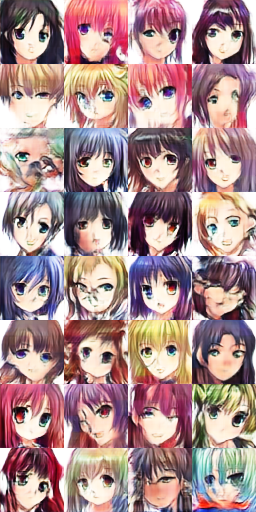}&
        		\includegraphics[width=0.25\linewidth]{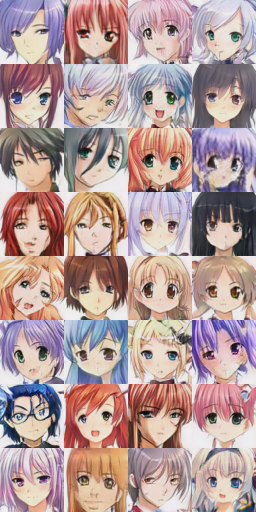} 
        		\\
	(a) WGAN-0GP & (b) MEG & (c) VERA & (d) EBM-BB~(ours) \\
        		
        	\end{tabular}
        	\caption{
        	Generated samples on ANIMEFACE with our method and other generative models
        	}
        	\label{anemiface}%
        \end{figure*}

        From \cref{image generation}, we see that our EBM-BB model is the best performing EBM on CIFAR-10, though it is surpassed by NCSN and DDPM. This is not surprising as NCSN and DDPM focus on sample quality rather than optimising a data likelihood. Further note that sampling NCSN and DDPM is significantly more expensive than our method. On ANIMEFACE our method is highly competitive. \cref{cifar10,anemiface} show samples from different models. For ANIMEFACE, our method generates more diverse samples in terms of face parts than the baselines. All models predominantly generate female faces. This suggests that while our approach captures more density modes than the baselines, all approaches still misses several modes. We further draw attention to several corrupt samples generated by VERA, despite extensive parameter search. For CIFAR-10, we observe no immediate differences in the generative capabilities between models.

        \paragraph{Capacity usage}
        For the proposal distribution $p_g$ to be adaptive, the associated generator network $G(\z)$ should be able to use as much of its available capacity as possible. To compare different methods, we consider an implicit measure of capacity usage $\mathcal{C}_{\z}$ that locally measures the intrinsic dimensionality of $G(\z)$. In particular, we use the anisotropy index \citep{wang2021geometry} that, for a given $\z$, measure the standard deviation of the norm of the directional derivative of $G(\z)$ along the input dimensions, i.e.
        \begin{align}
            \mathcal{C}_{\z} &= \mathrm{std}\left( \{ \| \mat{J}_{\z} \vec{e}_i \| \}_{i = 1\ldots d} \right),
        \end{align}
        \begin{wraptable}[7]{r}{0.5\linewidth}
            \vspace{-7mm}
            \caption{Anisotropy indices ($\downarrow$).}
            \label{disentangling}
            \resizebox{\linewidth}{!}{%
            \begin{tabular}{clclcl}
            \toprule
            \multicolumn{2}{c}{Model} & \multicolumn{2}{c}{CIFAR-10} & \multicolumn{2}{c}{ANIMEFACE} \\ \midrule
            \multicolumn{2}{c}{WGAN-0GP} & \multicolumn{2}{c}{1.0823 $\pm$ 0.005} & \multicolumn{2}{c}{2.4366 $\pm$ 0.129} \\
            \multicolumn{2}{c}{MEG} & \multicolumn{2}{c}{0.9467 $\pm$ 0.009} & \multicolumn{2}{c}{2.3600 $\pm$ 0.142} \\
            \multicolumn{2}{c}{VERA} & \multicolumn{2}{c}{3.9694 $\pm$ 0.055} & \multicolumn{2}{c}{2.2929 $\pm$ 0.2074} \\
            \multicolumn{2}{c}{EBM-0GP~(ours)} & \multicolumn{2}{c}{1.0688 $\pm$ 0.0226} & \multicolumn{2}{c}{3.4560 $\pm$ 0.0521} \\
            \multicolumn{2}{c}{EBM-BB~(ours)} & \multicolumn{2}{c}{\textbf{0.9431 $\pm$ 0.0191}} & \multicolumn{2}{c}{\textbf{1.8016 $\pm$ 0.1345}} \\ \bottomrule
            \end{tabular} }
        \end{wraptable}
        where $\mathrm{std}(\cdot)$ computes the standard deviation of the input, and $\vec{e}_i$ denotes the $i^{\mathrm{th}}$ standard basis vector. A small $\mathcal{C}_{\z}$ indicates that the different input dimensions contribute equally to the output of $G$, which imply good capacity usage. We measure the mean value of $\mathcal{C}_{\z}$ for $\z \sim \mathcal{N}(\vec{0}, \mat{I}_d)$ and report mean and standard deviation of the result running for several times in \cref{disentangling}. We observe that most models, with the exception of VERA, perform well on CIFAR-10, while on ANIMEFACE there is more diversity. In both cases, EBM-BB has the best capacity usage. The large difference between EBM-0GP and EBM-BB on ANIMEFACE suggest that our proposed upper bound helps increase the entropy of the proposal distribution $p_g$.

        \begin{table}[tb]
            \centering
            \caption{Comparison in terms of FID, Inception Score, $F_8$ score (weights recall higher than precision) and $F_{\sfrac{1}{8}}$ score (weights precision higher than recall).}
            \label{image generation}
            \begin{tabular}{clclllclcl}
            \toprule
            \multicolumn{2}{c}{Model} & \multicolumn{2}{c}{Inception$\uparrow$} & \multicolumn{2}{c}{FID$\downarrow$} & \multicolumn{2}{c}{$F_8\!\!\uparrow$} & \multicolumn{2}{c}{$F_{\sfrac{1}{8}}\!\uparrow$} \\ \midrule
            \multicolumn{10}{c}{CIFAR-10} \\ \midrule
            \multicolumn{2}{c}{DCGAN} & \multicolumn{2}{c}{6.64} & \multicolumn{2}{c}{49.03} & \multicolumn{2}{c}{0.795} & \multicolumn{2}{c}{0.83} \\
            \multicolumn{2}{c}{WGAN-0GP} & \multicolumn{2}{c}{7.24 $\pm$ 0.035} & \multicolumn{2}{c}{29.31 $\pm$ 0.185} & \multicolumn{2}{c}{0.92 $\pm$ 0.010} & \multicolumn{2}{c}{0.95 $\pm$ 0.010} \\
            \multicolumn{2}{c}{CoopNets~\citeyearpar{xie2018cooperative}} & \multicolumn{2}{c}{6.55} & \multicolumn{2}{c}{36.4} & \multicolumn{2}{c}{-} & \multicolumn{2}{c}{-} \\
            \multicolumn{2}{c}{CoopNets~\citeyearpar{xie2021cooperative}} & \multicolumn{2}{c}{-} & \multicolumn{2}{c}{33.61} & \multicolumn{2}{c}{-} & \multicolumn{2}{c}{-} \\
            \multicolumn{2}{c}{EBM-VAE} & \multicolumn{2}{c}{6.65} & \multicolumn{2}{c}{36.2} & \multicolumn{2}{c}{-} & \multicolumn{2}{c}{-} \\
            \multicolumn{2}{c}{IGEBM} & \multicolumn{2}{c}{6.78} & \multicolumn{2}{c}{38.2} & \multicolumn{2}{c}{-} & \multicolumn{2}{c}{-} \\
            \multicolumn{2}{c}{MEG} & \multicolumn{2}{c}{6.62 $\pm$ 0.243} & \multicolumn{2}{c}{34.55 $\pm$ 1.145} & \multicolumn{2}{c}{0.88 $\pm$ 0.001} & \multicolumn{2}{c}{0.92 $\pm$ 0.010} \\
            \multicolumn{2}{c}{VERA} & \multicolumn{2}{c}{5.06 $\pm$  0.555} & \multicolumn{2}{c}{66.38 $\pm$ 6.635} & \multicolumn{2}{c}{0.58 $\pm$ 0.080} & \multicolumn{2}{c}{0.79 $\pm$ 0.005} \\
            \multicolumn{2}{c}{NCSN} & \multicolumn{2}{c}{8.87 $\pm$  0.12} & \multicolumn{2}{c}{25.32} & \multicolumn{2}{c}{-} & \multicolumn{2}{c}{-} \\
            \multicolumn{2}{c}{DDPM} & \multicolumn{2}{c}{\textbf{9.03}} & \multicolumn{2}{c}{\textbf{7.76}} & \multicolumn{2}{c}{\textbf{0.98}} & \multicolumn{2}{c}{\textbf{0.99}} \\
            \multicolumn{2}{c}{EBM-0GP~(ours)} & \multicolumn{2}{c}{6.90 $\pm$ 0.032} & \multicolumn{2}{c}{35.42 $\pm$ 0.582} & \multicolumn{2}{c}{0.90 $\pm$ 0.004} & \multicolumn{2}{c}{0.93 $\pm$ 0.002} \\
            \multicolumn{2}{c}{EBM-BB~(ours)} & \multicolumn{2}{c}{7.45 $\pm$ 0.014} & \multicolumn{2}{c}{28.63 $\pm$ 0.290} & \multicolumn{2}{c}{0.93 $\pm$ 0.001} & \multicolumn{2}{c}{0.95 $\pm$ 0.008} \\ \midrule
            \multicolumn{10}{c}{ANIMEFACE} \\ \midrule
            \multicolumn{2}{c}{WGAN-0GP} & \multicolumn{2}{c}{2.22 $\pm$ 0.030} & \multicolumn{2}{c}{9.76 $\pm$ 0.674} & \multicolumn{2}{c}{\textbf{0.95 $\pm$ 0.005}} & \multicolumn{2}{c}{\textbf{0.98 $\pm$ 0.005}} \\
            \multicolumn{2}{c}{MEG} & \multicolumn{2}{c}{2.20 $\pm$ 0.020} & \multicolumn{2}{c}{9.31 $\pm$ 0.007} & \multicolumn{2}{c}{\textbf{0.95 $\pm$ 0.005}} & \multicolumn{2}{c}{\textbf{0.98 $\pm$ 0.001}} \\
            \multicolumn{2}{c}{VERA} & \multicolumn{2}{c}{2.15 $\pm $ 0.001} & \multicolumn{2}{c}{41.00 $\pm$ 1.072} & \multicolumn{2}{c}{0.515 $\pm$ 0.078} & \multicolumn{2}{c}{0.78 $\pm$ 0.013} \\
            \multicolumn{2}{c}{DDPM} & \multicolumn{2}{c}{2.18} & \multicolumn{2}{c}{\textbf{8.81}} & \multicolumn{2}{c}{0.94} & \multicolumn{2}{c}{\textbf{0.98}} \\
            \multicolumn{2}{c}{EBM-0GP~(ours)} & \multicolumn{2}{c}{\textbf{2.26 $\pm$ 0.017}} & \multicolumn{2}{c}{20.53 $\pm$ 0.524} & \multicolumn{2}{c}{0.889 $\pm$ 0.008} & \multicolumn{2}{c}{0.909 $\pm$ 0.019} \\
            \multicolumn{2}{c}{EBM-BB~(ours)} & \multicolumn{2}{c}{\textbf{2.26 $\pm$ 0.005}} & \multicolumn{2}{c}{12.75 $\pm$ 0.045} & \multicolumn{2}{c}{0.94 $\pm$ 0.001} & \multicolumn{2}{c}{0.96 $\pm$ 0.005} \\ \bottomrule
            \end{tabular}
        \end{table}

        \paragraph{Out-of-distribution detection}
        As EBMs are density estimators, they should, in principle, assign low likelihood to out-of-distribution (OOD) observations. OOD detection performance can then be seen as a proxy for the quality of the estimated density. To test how well our model fares in this regard, we follow previous work~\citep{hendrycks2016baseline, hendrycks2018deep, alemi2018uncertainty, choi2018waic, ren2019likelihood, havtorn} and report the threshold independent evaluation metrics of Area Under the Receiver Operator Characteristic (AUROC$\uparrow$), Area Under the Precision Recall Curve (AUPRC$\uparrow$) and False Positive Rate at 80\% (FPR80$\downarrow$), where the arrow indicates the direction of improvement of the metrics. The results are reported in \cref{OOD}. Each column of the table takes the form `In-distribution / Out-of-distribution' in reference to the training and test set, respectively.
        We observe that on CIFAR-10~/~SVHN, EBM-BB is the top performer. On  CIFAR-10~/~CIFAR-100 the overall performance degraded significantly, EBM-0GP fared noticeable better than other models. The overall degradation may be caused by the strong similarity between CIFAR-10 and CIFAR-100. For ANIMEFACE~/~Bedroom a similar situation occurs, even if these datasets are highly dissimilar, but we observe EBM-0GP is much better than other models.
        
        No clear winner can be found from this study. We note that our method consistently performs well, but surprisingly, so does WGAN-0GP even if it is not a density estimator. As OOD detection is a task that comes with many subtle pitfalls \citep{havtorn}, we suggest that these results should be taken with a grain of salt even if our model performs well.

        \begin{table}[tb]
            \centering
            \caption{AUROC$\uparrow$, AUPRC$\uparrow$ and FPR80$\downarrow$ for OOD detection for `train / test' datasets.}
            \label{OOD}
            \resizebox{\textwidth}{!}{%
            \begin{tabular}{cccccccccc}
            \toprule
            \multirow{2}{*}{Model} & \multicolumn{3}{c}{CIFAR-10~/~SVHN} & \multicolumn{3}{c}{CIFAR-10~/~CIFAR-100} & \multicolumn{3}{c}{ANIMEFACE~/~Bedroom} \\ \cmidrule(lr){2-4}\cmidrule(lr){5-7}\cmidrule(lr){8-10}%
             & AUROC$\uparrow$ & AUPRC$\uparrow$ & FPR80$\downarrow$ & AUROC$\uparrow$ & AUPRC$\uparrow$ & FPR80$\downarrow$ & AUROC$\uparrow$ & AUPRC$\uparrow$ & FPR80$\downarrow$ \\ \midrule
            WGAN-0GP & 0.8 & 0.83 & 0.39 & 0.54 & 0.55 & 0.77 & 0.51 & 0.48 & 0.72 \\
            MEG & 0.79 & 0.81 & 0.42 & 0.52 & 0.53 & 0.8 & 0.56 & 0.53 & 0.77 \\
            VERA & 0.62 & 0.64 & 0.64 & 0.51 & 0.51 & 0.79 & 0.60 & 0.525 & 0.6 \\
            EBM-0GP~(ours) & 0.66 & 0.69 & 0.67 & \textbf{0.64} & \textbf{0.63} & \textbf{0.64} & \textbf{0.67} & \textbf{0.63} & \textbf{0.53} \\
            EBM-BB~(ours) & \textbf{0.88} & \textbf{0.86} & \textbf{0.1997} & 0.53 & 0.52 & 0.7765 & 0.53 & 0.505 & 0.83 \\ \bottomrule
            \end{tabular}
            }
        \end{table}

        \paragraph{Running time}
        \begin{wraptable}[11]{r}{0.3\linewidth}
            \vspace{-2mm}
            \caption{Total training time.}
            \label{tbl:runtime}
            \vspace{2mm}
            \resizebox{\linewidth}{!}{%
            \begin{tabular}{clclcl}
            \toprule
            \multicolumn{2}{c}{Model} & \multicolumn{2}{l}{Iterations} & \multicolumn{2}{l}{Runtime} \\ \midrule
            \multicolumn{6}{c}{CIFAR-10} \\ \midrule
            \multicolumn{2}{c}{WGAN-0GP} & \multicolumn{2}{c}{100000} & \multicolumn{2}{c}{4h} \\
            \multicolumn{2}{c}{MEG} & \multicolumn{2}{c}{100000} & \multicolumn{2}{c}{4.5h} \\
            \multicolumn{2}{c}{VERA} & \multicolumn{2}{c}{200000} & \multicolumn{2}{c}{10h} \\
            \multicolumn{2}{c}{EBM-0GP} & \multicolumn{2}{c}{200000} & \multicolumn{2}{c}{15h} \\
            \multicolumn{2}{c}{EBM-BB} & \multicolumn{2}{c}{200000} & \multicolumn{2}{c}{20h} \\ \midrule
            \multicolumn{6}{c}{ANIMEFACE} \\ \midrule
            \multicolumn{2}{c}{WGAN-0GP} & \multicolumn{2}{c}{100000} & \multicolumn{2}{c}{13h} \\
            \multicolumn{2}{c}{MEG} & \multicolumn{2}{c}{100000} & \multicolumn{2}{c}{20h} \\
            \multicolumn{2}{c}{VERA} & \multicolumn{2}{c}{200000} & \multicolumn{2}{c}{44h} \\
            \multicolumn{2}{c}{EBM-0GP} & \multicolumn{2}{c}{100000} & \multicolumn{2}{c}{38h} \\
            \multicolumn{2}{c}{EBM-BB} & \multicolumn{2}{c}{100000} & \multicolumn{2}{c}{40h} \\ \bottomrule
            \end{tabular}
            }
        \end{wraptable}
        \Cref{tbl:runtime} gives an overview of the time required to train the proposed models and the baselines. All models were trained on a single 12GB Titan GPU. 
        We observe that our model should be trained with a lower learning rate, and therefore may increase the total number of epochs.

    \section{Discussion}
    The first observation in our work is that current methods for training energy-based models (EBMs) interchangeably minimize and maximize a lower bound. As this may be a potential source of training instability, we propose to bound the negative log-likelihood from above and below and switch between bounds when minimizing and maximizing. The lower bound, $\floor{\L}$, is similar to existing ones, but we provide a new algorithm for its realization. Unlike past work, this algorithm does not need additional networks and does not rely on hard-to-tune parameters; our only parameter controls the number of iterations for the Jacobian approximation and represents a trade-off between the accuracy of the bound and computational budget. The upper bound, $\ceil{\L}$, is new to the literature but similar to the common regularization practice of introducing gradient penalties. To the best of our knowledge, this is the first time that gradient penalties have been derived from the perspective of bounding the log-likelihood. It is rewarding that current better engineering practice can be justified from a probabilistic perspective.
    Empirically, we find our proposed model generally performs as well or better than some of the current state-of-the-art on a variety of tasks. The evidence suggests that our bidirectional bounds allow the generator to increase its entropy and capacity usage. We see this both directly and through improved sample quality on diverse datasets.
    
    \paragraph{Limitations}
    The current drawbacks of the method are mainly three concerns. First, we find that a smaller-than-usual learning rate helps our model. While we expect this to be a matter of implementation rather than a more profound concern, the current implication is that training is approximately twice that of a EBM baseline. Second, our bounds rely on computing the smallest singular value of the Jacobian of the generator. We provide an efficient implementation of a method for this task, but it needs to run for several iterations to guarantee convergence. In practice, we stop after a fixed, low number of iterations, which we found to work well, but technically this violates the bound. This seems to be an unavoidable aspect of our approach, but we find that the benefits significantly compensate for this issue. The availability of a fast (approximate) entropy estimator that does not require additional networks and parameters is highly valuable when training EBMs. Finally, for the upper bound, we need an estimate of the volume of support $M$ of the proposal distribution $p_g$. We do not have a viable method for this, and, in practice, we treat $M$ as a hyper-parameter. We have not found it difficult to tune this parameter, but it nonetheless constitutes a limitation.

    \paragraph{Negative societal impact}
    All high-capacity generative models carry the risk of being used for misinformation, and our model is no exception. The value of EBMs over GANs is that they come with a likelihood function, which is more valuable in data analysis, than e.g.\@ for creating deepfakes.

\begin{ack}
This work was funded in part by the Novo Nordisk Foundation through the Center for Basic Machine Learning Research in Life Science (NNF20OC0062606). It also received funding from the European Research Council (ERC) under the European Union’s Horizon 2020 research, innovation programme (757360), National science foundation of China under grant 61771305 and Shanghai Municipal Science and Technology Major Project (2021SHZDZX0102). JF was supported in part by the Novo Nordisk Foundation (NNF20OC0065611) and the Independent Research Fund Denmark (9131-00082B). SH was supported in part by a research grant (15334) from VILLUM FONDEN. The authors are grateful to anonymous reviewers and the handling area chair for valuable discussions and feedback on an early version of this manuscript.
\end{ack}

\newpage

\clearpage
\bibliography{bibliography}

\begin{thebibliography}{65}
\providecommand{\natexlab}[1]{#1}
\providecommand{\url}[1]{\texttt{#1}}
\expandafter\ifx\csname urlstyle\endcsname\relax
  \providecommand{\doi}[1]{doi: #1}\else
  \providecommand{\doi}{doi: \begingroup \urlstyle{rm}\Url}\fi

\bibitem[Abbasnejad et~al.(2019)Abbasnejad, Shi, Hengel, and
  Liu]{abbasnejad2019generative}
M.~E. Abbasnejad, Q.~Shi, A.~v.~d. Hengel, and L.~Liu.
\newblock A generative adversarial density estimator.
\newblock In \emph{Proceedings of the IEEE/CVF Conference on Computer Vision
  and Pattern Recognition}, pages 10782--10791, 2019.

\bibitem[Abbasnejad et~al.(2020)Abbasnejad, Shi, van~den Hengel, and
  Liu]{abbasnejad2020gade}
M.~E. Abbasnejad, J.~Shi, A.~van~den Hengel, and L.~Liu.
\newblock Gade: A generative adversarial approach to density estimation and its
  applications.
\newblock \emph{International Journal of Computer Vision}, 128\penalty0
  (10):\penalty0 2731--2743, 2020.

\bibitem[Ackley et~al.(1985)Ackley, Hinton, and Sejnowski]{hinton1985learning}
D.~H. Ackley, G.~E. Hinton, and T.~J. Sejnowski.
\newblock A learning algorithm for boltzmann machines.
\newblock \emph{Cognitive Science}, 9\penalty0 (1):\penalty0 147--169, 1985.

\bibitem[Alemi et~al.(2018)Alemi, Fischer, and Dillon]{alemi2018uncertainty}
A.~A. Alemi, I.~Fischer, and J.~V. Dillon.
\newblock Uncertainty in the variational information bottleneck.
\newblock \emph{arXiv preprint arXiv:1807.00906}, 2018.

\bibitem[Arjovsky et~al.(2017)Arjovsky, Chintala, and
  Bottou]{arjovsky2017wasserstein}
M.~Arjovsky, S.~Chintala, and L.~Bottou.
\newblock Wasserstein generative adversarial networks.
\newblock In \emph{International conference on machine learning}, pages
  214--223. PMLR, 2017.

\bibitem[Burda et~al.(2015)Burda, Grosse, and Salakhutdinov]{pmlr-v38-burda15}
Y.~Burda, R.~Grosse, and R.~Salakhutdinov.
\newblock {Accurate and conservative estimates of MRF log-likelihood using
  reverse annealing}.
\newblock In G.~Lebanon and S.~V.~N. Vishwanathan, editors, \emph{Proceedings
  of the Eighteenth International Conference on Artificial Intelligence and
  Statistics}, volume~38 of \emph{Proceedings of Machine Learning Research},
  pages 102--110, San Diego, California, USA, 09--12 May 2015. PMLR.

\bibitem[Che et~al.(2020)Che, Zhang, Sohl-Dickstein, Larochelle, Paull, Cao,
  and Bengio]{che2020your}
T.~Che, R.~Zhang, J.~Sohl-Dickstein, H.~Larochelle, L.~Paull, Y.~Cao, and
  Y.~Bengio.
\newblock Your {GAN} is secretly an energy-based model and you should use
  discriminator driven latent sampling.
\newblock In H.~Larochelle, M.~Ranzato, R.~Hadsell, M.~F. Balcan, and H.~Lin,
  editors, \emph{Advances in Neural Information Processing Systems}, volume~33,
  pages 12275--12287. Curran Associates, Inc., 2020.

\bibitem[Choi et~al.(2018)Choi, Jang, and Alemi]{choi2018waic}
H.~Choi, E.~Jang, and A.~A. Alemi.
\newblock {WAIC}, but why? {G}enerative ensembles for robust anomaly detection.
\newblock \emph{arXiv preprint arXiv:1810.01392}, 2018.

\bibitem[Dai et~al.(2017)Dai, Almahairi, Bachman, Hovy, and
  Courville]{dai2017calibrating}
Z.~Dai, A.~Almahairi, P.~Bachman, E.~Hovy, and A.~Courville.
\newblock Calibrating energy-based generative adversarial networks.
\newblock In \emph{International Conference on Learning Representations}, 2017.

\bibitem[Du and Mordatch(2019)]{du2019implicit}
Y.~Du and I.~Mordatch.
\newblock Implicit generation and modeling with energy based models.
\newblock In H.~Wallach, H.~Larochelle, A.~Beygelzimer, F.~d\textquotesingle
  Alch\'{e}-Buc, E.~Fox, and R.~Garnett, editors, \emph{Advances in Neural
  Information Processing Systems}, volume~32. Curran Associates, Inc., 2019.

\bibitem[Frellsen et~al.(2016)Frellsen, Winther, Ghahramani, and
  Ferkinghoff-Borg]{pmlr-v51-frellsen16}
J.~Frellsen, O.~Winther, Z.~Ghahramani, and J.~Ferkinghoff-Borg.
\newblock Bayesian generalised ensemble markov chain monte carlo.
\newblock In A.~Gretton and C.~C. Robert, editors, \emph{Proceedings of the
  19th International Conference on Artificial Intelligence and Statistics},
  volume~51 of \emph{Proceedings of Machine Learning Research}, pages 408--416,
  Cadiz, Spain, 09--11 May 2016. PMLR.

\bibitem[Grathwohl et~al.(2021)Grathwohl, Kelly, Hashemi, Norouzi, Swersky, and
  Duvenaud]{nomcmc}
W.~S. Grathwohl, J.~J. Kelly, M.~Hashemi, M.~Norouzi, K.~Swersky, and
  D.~Duvenaud.
\newblock No {MCMC} for me: Amortized sampling for fast and stable training of
  energy-based models.
\newblock In \emph{International Conference on Learning Representations}, 2021.

\bibitem[Grosse et~al.(2013)Grosse, Maddison, and
  Salakhutdinov]{NIPS2013_fb60d411}
R.~B. Grosse, C.~J. Maddison, and R.~R. Salakhutdinov.
\newblock Annealing between distributions by averaging moments.
\newblock In C.~J.~C. Burges, L.~Bottou, M.~Welling, Z.~Ghahramani, and K.~Q.
  Weinberger, editors, \emph{Advances in Neural Information Processing
  Systems}, volume~26. Curran Associates, Inc., 2013.

\bibitem[Gulrajani et~al.(2017)Gulrajani, Ahmed, Arjovsky, Dumoulin, and
  Courville]{gulrajani2017improved}
I.~Gulrajani, F.~Ahmed, M.~Arjovsky, V.~Dumoulin, and A.~C. Courville.
\newblock Improved training of wasserstein {GAN}s.
\newblock In I.~Guyon, U.~V. Luxburg, S.~Bengio, H.~Wallach, R.~Fergus,
  S.~Vishwanathan, and R.~Garnett, editors, \emph{Advances in Neural
  Information Processing Systems}, volume~30. Curran Associates, Inc., 2017.

\bibitem[Gutmann and Hyv{\"a}rinen(2010)]{gutmann2010noise}
M.~Gutmann and A.~Hyv{\"a}rinen.
\newblock Noise-contrastive estimation: A new estimation principle for
  unnormalized statistical models.
\newblock In \emph{Proceedings of the Thirteenth International Conference on
  Artificial Intelligence and Statistics}, pages 297--304. JMLR Workshop and
  Conference Proceedings, 2010.

\bibitem[Han et~al.(2019)Han, Nijkamp, Fang, Hill, Zhu, and
  Wu]{han2019divergence}
T.~Han, E.~Nijkamp, X.~Fang, M.~Hill, S.-C. Zhu, and Y.~N. Wu.
\newblock Divergence triangle for joint training of generator model,
  energy-based model, and inferential model.
\newblock In \emph{Proceedings of the IEEE/CVF Conference on Computer Vision
  and Pattern Recognition}, pages 8670--8679, 2019.

\bibitem[Havtorn et~al.(2021)Havtorn, Frellsen, Hauberg, and
  Maal{\o}e]{havtorn}
J.~D. Havtorn, J.~Frellsen, S.~Hauberg, and L.~Maal{\o}e.
\newblock Hierarchical {VAE}s know what they don’t know.
\newblock In M.~Meila and T.~Zhang, editors, \emph{Proceedings of the 38th
  International Conference on Machine Learning}, volume 139 of
  \emph{Proceedings of Machine Learning Research}, pages 4117--4128. PMLR,
  18--24 Jul 2021.

\bibitem[Hendrycks and Gimpel(2017)]{hendrycks2016baseline}
D.~Hendrycks and K.~Gimpel.
\newblock A baseline for detecting misclassified and out-of-distribution
  examples in neural networks.
\newblock In \emph{International Conference on Learning Representations}, 2017.

\bibitem[Hendrycks et~al.(2019)Hendrycks, Mazeika, and
  Dietterich]{hendrycks2018deep}
D.~Hendrycks, M.~Mazeika, and T.~Dietterich.
\newblock Deep anomaly detection with outlier exposure.
\newblock In \emph{International Conference on Learning Representations}, 2019.

\bibitem[Hinton(2002)]{hinton2002training}
G.~E. Hinton.
\newblock Training products of experts by minimizing contrastive divergence.
\newblock \emph{Neural computation}, 14\penalty0 (8):\penalty0 1771--1800,
  2002.

\bibitem[Hinton and Sejnowski(1983)]{hinton1983optimal}
G.~E. Hinton and T.~J. Sejnowski.
\newblock Optimal perceptual inference.
\newblock In \emph{Proceedings of the IEEE conference on Computer Vision and
  Pattern Recognition}, volume 448, 1983.

\bibitem[Hinton et~al.(2006)Hinton, Osindero, and Teh]{hinton2006fast}
G.~E. Hinton, S.~Osindero, and Y.-W. Teh.
\newblock A fast learning algorithm for deep belief nets.
\newblock \emph{Neural computation}, 18\penalty0 (7):\penalty0 1527--1554,
  2006.

\bibitem[Ho et~al.(2020)Ho, Jain, and Abbeel]{ho2020denoising}
J.~Ho, A.~Jain, and P.~Abbeel.
\newblock Denoising diffusion probabilistic models.
\newblock In H.~Larochelle, M.~Ranzato, R.~Hadsell, M.~F. Balcan, and H.~Lin,
  editors, \emph{Advances in Neural Information Processing Systems}, volume~33,
  pages 6840--6851. Curran Associates, Inc., 2020.
\newblock URL
  \url{https://proceedings.neurips.cc/paper/2020/file/4c5bcfec8584af0d967f1ab10179ca4b-Paper.pdf}.

\bibitem[Hopfield(1982)]{Hopfield2554}
J.~J. Hopfield.
\newblock Neural networks and physical systems with emergent collective
  computational abilities.
\newblock \emph{Proceedings of the National Academy of Sciences}, 79\penalty0
  (8):\penalty0 2554--2558, 1982.
\newblock ISSN 0027-8424.

\bibitem[Hutchinson(1989)]{hutchinson1989stochastic}
M.~F. Hutchinson.
\newblock A stochastic estimator of the trace of the influence matrix for
  laplacian smoothing splines.
\newblock \emph{Communications in Statistics-Simulation and Computation},
  18\penalty0 (3):\penalty0 1059--1076, 1989.

\bibitem[Hyv{\"a}rinen(2005)]{hyvarinen2005estimation}
A.~Hyv{\"a}rinen.
\newblock Estimation of non-normalized statistical models by score matching.
\newblock \emph{Journal of Machine Learning Research}, 6\penalty0 (4), 2005.

\bibitem[Kang and Park(2020)]{kang2020ContraGAN}
M.~Kang and J.~Park.
\newblock {ContraGAN}: Contrastive learning for conditional image generation.
\newblock In H.~Larochelle, M.~Ranzato, R.~Hadsell, M.~F. Balcan, and H.~Lin,
  editors, \emph{Advances in Neural Information Processing Systems}, volume~33,
  pages 21357--21369. Curran Associates, Inc., 2020.

\bibitem[Kim and Bengio(2016)]{kim2016deep}
T.~Kim and Y.~Bengio.
\newblock Deep directed generative models with energy-based probability
  estimation.
\newblock \emph{arXiv preprint arXiv:1606.03439}, 2016.

\bibitem[Knyazev(1998)]{knyazev1998preconditioned}
A.~V. Knyazev.
\newblock Preconditioned eigensolvers—an oxymoron.
\newblock \emph{Electron. Trans. Numer. Anal}, 7:\penalty0 104--123, 1998.

\bibitem[Knyazev(2001)]{knyazev2001toward}
A.~V. Knyazev.
\newblock Toward the optimal preconditioned eigensolver: Locally optimal block
  preconditioned conjugate gradient method.
\newblock \emph{SIAM journal on scientific computing}, 23\penalty0
  (2):\penalty0 517--541, 2001.

\bibitem[Krizhevsky et~al.(2009)Krizhevsky, Hinton,
  et~al.]{krizhevsky2009learning}
A.~Krizhevsky, G.~Hinton, et~al.
\newblock Learning multiple layers of features from tiny images.
\newblock Technical report, University of Toronto, 2009.

\bibitem[Kumar et~al.(2020)Kumar, Poole, and Murphy]{kumar2020regularized}
A.~Kumar, B.~Poole, and K.~Murphy.
\newblock Regularized autoencoders via relaxed injective probability flow.
\newblock In \emph{International Conference on Artificial Intelligence and
  Statistics}, pages 4292--4301. PMLR, 2020.

\bibitem[Kumar et~al.(2019)Kumar, Ozair, Goyal, Courville, and
  Bengio]{kumar2019maximum}
R.~Kumar, S.~Ozair, A.~Goyal, A.~Courville, and Y.~Bengio.
\newblock Maximum entropy generators for energy-based models.
\newblock \emph{arXiv preprint arXiv:1901.08508}, 2019.

\bibitem[LeCun(1998)]{lecun1998mnist}
Y.~LeCun.
\newblock The {MNIST} database of handwritten digits.
\newblock \emph{http://yann. lecun. com/exdb/mnist/}, 1998.

\bibitem[LeCun et~al.(2006)LeCun, Chopra, Hadsell, Ranzato, and
  Huang]{lecun2006tutorial}
Y.~LeCun, S.~Chopra, R.~Hadsell, M.~Ranzato, and F.~Huang.
\newblock A tutorial on energy-based learning.
\newblock \emph{Predicting structured data}, 1\penalty0 (0), 2006.

\bibitem[Li and Turner(2018)]{li2017gradient}
Y.~Li and R.~E. Turner.
\newblock Gradient estimators for implicit models.
\newblock In \emph{International Conference on Learning Representations}, 2018.

\bibitem[Metropolis et~al.(1953)Metropolis, Rosenbluth, Rosenbluth, Teller, and
  Teller]{metropolis1953equation}
N.~Metropolis, A.~W. Rosenbluth, M.~N. Rosenbluth, A.~H. Teller, and E.~Teller.
\newblock Equation of state calculations by fast computing machines.
\newblock \emph{The journal of chemical physics}, 21\penalty0 (6):\penalty0
  1087--1092, 1953.

\bibitem[Miyato et~al.(2018)Miyato, Kataoka, Koyama, and
  Yoshida]{miyato2018spectral}
T.~Miyato, T.~Kataoka, M.~Koyama, and Y.~Yoshida.
\newblock Spectral normalization for generative adversarial networks.
\newblock In \emph{International Conference on Learning Representations}, 2018.

\bibitem[Neal et~al.(2011)]{neal2011mcmc}
R.~M. Neal et~al.
\newblock {MCMC} using hamiltonian dynamics.
\newblock \emph{Handbook of markov chain monte carlo}, 2\penalty0
  (11):\penalty0 2, 2011.

\bibitem[Neudecker(1992)]{neudecker1992matrix}
H.~Neudecker.
\newblock A matrix trace inequality.
\newblock \emph{Journal of mathematical analysis and applications},
  166\penalty0 (1):\penalty0 302--303, 1992.

\bibitem[Osogami(2017)]{osogami2017boltzmann}
T.~Osogami.
\newblock Boltzmann machines and energy-based models.
\newblock \emph{arXiv preprint arXiv:1708.06008}, 2017.

\bibitem[Paszke et~al.(2017)Paszke, Gross, Chintala, Chanan, Yang, DeVito, Lin,
  Desmaison, Antiga, and Lerer]{paszke2017automatic}
A.~Paszke, S.~Gross, S.~Chintala, G.~Chanan, E.~Yang, Z.~DeVito, Z.~Lin,
  A.~Desmaison, L.~Antiga, and A.~Lerer.
\newblock Automatic differentiation in pytorch.
\newblock \emph{NIPS 2017 Workshop Autodiff}, 2017.

\bibitem[Radford et~al.(2016)Radford, Metz, and
  Chintala]{radford2015unsupervised}
A.~Radford, L.~Metz, and S.~Chintala.
\newblock Unsupervised representation learning with deep convolutional
  generative adversarial networks.
\newblock In \emph{International Conference on Learning Representations}, 2016.

\bibitem[Ren et~al.(2019)Ren, Liu, Fertig, Snoek, Poplin, Depristo, Dillon, and
  Lakshminarayanan]{ren2019likelihood}
J.~Ren, P.~J. Liu, E.~Fertig, J.~Snoek, R.~Poplin, M.~Depristo, J.~Dillon, and
  B.~Lakshminarayanan.
\newblock Likelihood ratios for out-of-distribution detection.
\newblock In H.~Wallach, H.~Larochelle, A.~Beygelzimer, F.~d\textquotesingle
  Alch\'{e}-Buc, E.~Fox, and R.~Garnett, editors, \emph{Advances in Neural
  Information Processing Systems}, volume~32. Curran Associates, Inc., 2019.

\bibitem[Sajjadi et~al.(2018)Sajjadi, Bachem, Lu{\v c}i{\'c}, Bousquet, and
  Gelly]{precision_recall_distributions}
M.~S.~M. Sajjadi, O.~Bachem, M.~Lu{\v c}i{\'c}, O.~Bousquet, and S.~Gelly.
\newblock {Assessing Generative Models via Precision and Recall}.
\newblock In \emph{{Advances in Neural Information Processing Systems
  (NeurIPS)}}, 2018.

\bibitem[Salakhutdinov and Hinton(2009)]{pmlr-v5-salakhutdinov09a}
R.~Salakhutdinov and G.~Hinton.
\newblock Deep boltzmann machines.
\newblock In D.~van Dyk and M.~Welling, editors, \emph{Proceedings of the
  Twelth International Conference on Artificial Intelligence and Statistics},
  volume~5 of \emph{Proceedings of Machine Learning Research}, pages 448--455,
  Hilton Clearwater Beach Resort, Clearwater Beach, Florida USA, 16--18 Apr
  2009. PMLR.

\bibitem[Salakhutdinov and Murray(2008)]{salakhutdinov:icml08a}
R.~Salakhutdinov and I.~Murray.
\newblock On the quantitative analysis of deep belief networks.
\newblock In A.~McCallum and S.~Roweis, editors, \emph{Proceedings of the 25th
  Annual International Conference on Machine Learning (ICML 2008)}, pages
  872--879. Omnipress, 2008.

\bibitem[Scellier(2020)]{scellier2020deep}
B.~Scellier.
\newblock \emph{A deep learning theory for neural networks grounded in
  physics}.
\newblock PhD thesis, Université de Montréal, Quebec, Canada, 2020.

\bibitem[Shi et~al.(2018)Shi, Sun, and Zhu]{shi2018spectral}
J.~Shi, S.~Sun, and J.~Zhu.
\newblock A spectral approach to gradient estimation for implicit
  distributions.
\newblock In \emph{International Conference on Machine Learning}, pages
  4644--4653. PMLR, 2018.

\bibitem[Smolensky(1986)]{10.5555/104279.104290}
P.~Smolensky.
\newblock \emph{Information Processing in Dynamical Systems: Foundations of
  Harmony Theory}, page 194–281.
\newblock MIT Press, Cambridge, MA, USA, 1986.

\bibitem[Song and Ermon(2019)]{song2019generative}
Y.~Song and S.~Ermon.
\newblock Generative modeling by estimating gradients of the data distribution.
\newblock In H.~Wallach, H.~Larochelle, A.~Beygelzimer, F.~d\textquotesingle
  Alch\'{e}-Buc, E.~Fox, and R.~Garnett, editors, \emph{Advances in Neural
  Information Processing Systems}, volume~32. Curran Associates, Inc., 2019.

\bibitem[Thanh-Tung et~al.(2019)Thanh-Tung, Tran, and
  Venkatesh]{thanh2019improving}
H.~Thanh-Tung, T.~Tran, and S.~Venkatesh.
\newblock Improving generalization and stability of generative adversarial
  networks.
\newblock In \emph{International Conference on Learning Representations}, 2019.

\bibitem[Wang and Ponce(2021)]{wang2021geometry}
B.~Wang and C.~R. Ponce.
\newblock The geometry of deep generative image models and its applications.
\newblock \emph{arXiv preprint arXiv:2101.06006}, 2021.

\bibitem[Wu et~al.(2018)Wu, Xie, Lu, and Zhu]{wu2018sparse}
Y.~N. Wu, J.~Xie, Y.~Lu, and S.-C. Zhu.
\newblock Sparse and deep generalizations of the frame model.
\newblock \emph{Annals of Mathematical Sciences and Applications}, 3\penalty0
  (1):\penalty0 211--254, 2018.

\bibitem[Xie et~al.(2015)Xie, Hu, Zhu, and Wu]{xie2015learning}
J.~Xie, W.~Hu, S.-C. Zhu, and Y.~N. Wu.
\newblock Learning sparse {FRAME} models for natural image patterns.
\newblock \emph{International Journal of Computer Vision}, 114\penalty0
  (2):\penalty0 91--112, 2015.

\bibitem[Xie et~al.(2016)Xie, Lu, Zhu, and Wu]{xie2016inducing}
J.~Xie, Y.~Lu, S.-C. Zhu, and Y.~N. Wu.
\newblock Inducing wavelets into random fields via generative boosting.
\newblock \emph{Applied and Computational Harmonic Analysis}, 41\penalty0
  (1):\penalty0 4--25, 2016.

\bibitem[Xie et~al.(2017)Xie, Zhu, and Nian~Wu]{xie2017synthesizing}
J.~Xie, S.-C. Zhu, and Y.~Nian~Wu.
\newblock Synthesizing dynamic patterns by spatial-temporal generative convnet.
\newblock In \emph{Proceedings of the ieee conference on computer vision and
  pattern recognition}, pages 7093--7101, 2017.

\bibitem[Xie et~al.(2018{\natexlab{a}})Xie, Lu, Gao, and
  Wu]{xie2018cooperative}
J.~Xie, Y.~Lu, R.~Gao, and Y.~N. Wu.
\newblock Cooperative learning of energy-based model and latent variable model
  via {MCMC} teaching.
\newblock In \emph{Proceedings of the AAAI Conference on Artificial
  Intelligence}, volume~32, 2018{\natexlab{a}}.

\bibitem[Xie et~al.(2018{\natexlab{b}})Xie, Lu, Gao, Zhu, and
  Wu]{xie2018cooperative2}
J.~Xie, Y.~Lu, R.~Gao, S.-C. Zhu, and Y.~N. Wu.
\newblock Cooperative training of descriptor and generator networks.
\newblock \emph{IEEE transactions on pattern analysis and machine
  intelligence}, 42\penalty0 (1):\penalty0 27--45, 2018{\natexlab{b}}.

\bibitem[Xie et~al.(2018{\natexlab{c}})Xie, Zheng, Gao, Wang, Zhu, and
  Wu]{xie2018learning}
J.~Xie, Z.~Zheng, R.~Gao, W.~Wang, S.-C. Zhu, and Y.~N. Wu.
\newblock Learning descriptor networks for 3d shape synthesis and analysis.
\newblock In \emph{Proceedings of the IEEE conference on computer vision and
  pattern recognition}, pages 8629--8638, 2018{\natexlab{c}}.

\bibitem[Xie et~al.(2021{\natexlab{a}})Xie, Zheng, Fang, Zhu, and
  Wu]{xie2021cooperative}
J.~Xie, Z.~Zheng, X.~Fang, S.-C. Zhu, and Y.~N. Wu.
\newblock Cooperative training of fast thinking initializer and slow thinking
  solver for conditional learning.
\newblock \emph{IEEE Transactions on Pattern Analysis and Machine
  Intelligence}, 2021{\natexlab{a}}.

\bibitem[Xie et~al.(2021{\natexlab{b}})Xie, Zheng, and Li]{xie2021learning}
J.~Xie, Z.~Zheng, and P.~Li.
\newblock Learning energybased model with variational auto-encoder as amortized
  sampler.
\newblock In \emph{The Thirty-Fifth AAAI Conference on Artificial Intelligence
  (AAAI)}, volume~2, 2021{\natexlab{b}}.

\bibitem[Zhai et~al.(2016)Zhai, Cheng, Feris, and Zhang]{zhai2016generative}
S.~Zhai, Y.~Cheng, R.~Feris, and Z.~Zhang.
\newblock Generative adversarial networks as variational training of energy
  based models.
\newblock \emph{arXiv preprint arXiv:1611.01799}, 2016.

\bibitem[Zhu and Mumford(1998)]{zhu1998grade}
S.~C. Zhu and D.~Mumford.
\newblock Grade: {G}ibbs reaction and diffusion equations.
\newblock In \emph{Sixth International Conference on Computer Vision (IEEE Cat.
  No. 98CH36271)}, pages 847--854. IEEE, 1998.

\bibitem[Zhu et~al.(1998)Zhu, Wu, and Mumford]{zhu1998filters}
S.~C. Zhu, Y.~Wu, and D.~Mumford.
\newblock Filters, random fields and maximum entropy ({FRAME}): Towards a
  unified theory for texture modeling.
\newblock \emph{International Journal of Computer Vision}, 27\penalty0
  (2):\penalty0 107--126, 1998.

\end{thebibliography}
\clearpage

\maketitle

\appendix
\section{Theoretical  part}
\subsection{Proof of \textbf{Theorem~1} }
\begin{proof}
\begin{equation}
\begin{aligned}
 &\log \left[\expt_{\x \sim g(\x)} \left[f(x)\right]\right]-\expt_{\x \sim g(\x)} \left[\log f(x)\right]=\log \int f(x)g(x)dx-\int (\log f(x)) g(x)dx\\
 &=\log \left[f(\eta)\int g(x)dx\right]-\int (\log f(x))g(x)dx=\log f(\eta)\int g(x)dx-\int (\log f(x))g(x)dx\\
 &=\int\left[\log f(\eta)-\log f(x)\right]g(x)dx=\int g(x)\int_0^1\frac{d\left[\log f(t\eta+(1-t)x)\right]}{dt}dtdx\\
 &=\int g(x)\int_0^1 \nabla_{\tilde{x}} \log f(\tilde{x})(\eta-x)dtdx\\
 & \leq \int g(x)\left(\int_0^1 \vert \nabla_{\tilde{x}}\log f(\tilde{x})\vert^p dt\right)^{\frac{1}{p}}\left(\int_0^1\vert \eta-x \vert^q dt\right)^{\frac{1}{q}}dx \\
 &\leq \int g(x)\vert \eta-x \vert \left(\int_0^1 \vert \nabla_{\tilde{x}}\log f(\tilde{x})\vert^p dt\right)^{\frac{1}{p}}dx\\
 &\leq \left(\int g(x)\int_0^1\vert \nabla_{\tilde{x}} \log f(\tilde{x})\vert^p dtdx\right)^{\frac{1}{p}}\left(\int g(x)\vert \eta-x\vert^q dx\right)^{\frac{1}{q}},
 \end{aligned}
 \end{equation}
where $\tilde{x}=t\eta+(1-t)x$. The second equation is derived by mean value theorem for definite integrals. The first inequality is derived by Holder's inequality, so $p,q \geq 1$ and $\frac{1}{p}+\frac{1}{q}=1$. Because $g(x)$ has finite support, there exists an $M\geq0$ that satisfying: $\vert \eta-x \vert \leq M$, then we can get:
 \begin{equation}
 \begin{aligned}
\log \left[\expt_{\x \sim g(\x)} \left[f(x)\right]\right]-\expt_{\x \sim g(\x)} \left[\log f(x)\right] &\leq M \left(\int g(x)\int_0^1\vert \nabla_{\tilde{x}} \log f(\tilde{x})\vert^p dtdx\right)^{\frac{1}{p}}\\
&\leq M \left(\int g(x)\vert \nabla_{\hat{x}} \log f(\hat{x})\vert^p dx\right)^{\frac{1}{p}},
\end{aligned}
 \end{equation}
where $\hat{x}=t_0\eta+(1-t_0)x$ for a $t_0$~($0 \leq t_0 \leq 1$) using the mean value theorem.  Because $f(x)$ is L-Lipschitz continuous, then $\log f(x)$ is also Lipschitz continuous, so there exists an $m \geq 0$ satisfying 
 \begin{equation}
     \vert \nabla_{\hat{x}} \log f(\hat{x})\vert^p \leq \vert \nabla_{x} \log f(x)\vert^p +m, for \forall x,
 \end{equation}
 so we can get 
 \begin{equation}
 \begin{aligned}
     \log \left[\expt_{\x \sim g(\x)} \left[f(x)\right]\right]-\expt_{\x \sim g(\x)} \left[\log f(x)\right] &\leq M \left(\int g(x)\vert \nabla_{x} \log f(x)\vert^p dx+m\right)^{\frac{1}{p}}\\
     &\leq M (\expt_{\x \sim g(\x)} \left[\vert\nabla_x\log f(x)\vert^p\right]+m)^{\frac{1}{p}}.
\label{ori}
\end{aligned}
 \end{equation}
\end{proof}

\subsection{Proof of \textbf{Eq.~(18)}}
\begin{proof}
For a vector $u\in R^{1\times n}$ and a matrix $J\in R^{n\times m}$, we have
\begin{equation}
\begin{aligned}
 \vert uJ\vert_2^2&=uJJ^Tu^T=\operatorname{tr}(JJ^Tu^Tu)\\
 &\geq \lambda(JJ^T) \operatorname{tr}(u^Tu),
\end{aligned}
\end{equation}
where $\lambda$ is the smallest eigenvalue of $JJ^T$. The inequality holds because $JJ^T$ is a real symmetric matrix and $u^Tu$ is positive semidefinite~\citep{neudecker1992matrix}. Then we set $J=\J_{\z}$,$u= \nabla_{\x} E_{\theta} (G(\z)) + \nabla_{G(\z)}\log p_g(G(\z))$, we can obtain:
\begin{equation}
\begin{aligned}
    &\vert \nabla_{\x}E_{\theta} (G(\z))\J_{\z}
           + \nabla_{G(\z)}\log p_g(G(\z))\J_{\z}\vert_2^2 \geq \lambda(\J_{\z}\J_{\z}^T) \operatorname{tr}(u^Tu)\\
&= \lambda(\J_{\z}\J_{\z}^T)\vert u\vert_2^2=\lambda(\J_{\z}\J_{\z}^T)\vert \nabla_{\x} E_{\theta} (G(\z)) + \nabla_{G(\z)}\log p_g(G(\z))\vert_2^2
\end{aligned}
\end{equation}
Because $\lambda(\J_{\z}\J_{\z}^T)$ is the square of the smallest singular value of $\J_{\z}$, which we represent by $s_1$. So we can obtain:
\begin{equation}
    \vert \nabla_{\x} E_{\theta} (G(\z)) + \nabla_{G(\z)}\log p_g(G(\z))\vert_2 \leq \frac{\vert \nabla_{\x}E_{\theta} (G(\z))\J_{\z}
           + \nabla_{G(\z)}\log p_g(G(\z))\J_{\z}\vert_2}{s_1}.
\end{equation}
This proves Eq.~18.
\end{proof}

\section{Model Architecture}
In order apply the change-of-variables formula to get a density for the generator, we assume that $G: \mathbb{R}^d \rightarrow \mathbb{R}^D$ spans an immersed $d$-dimensional manifold in $\mathbb{R}^D$. This assumption place some restrictions on the architecture of the generator neural network.

The governing assumption is that the Jacobian of $G$ exist and has full rank. Existence is ensured as long as the chosen activation functions have at least one derivative almost everywhere. Smooth activations naturally satisfy this assumption, but it is worth noting that e.g.\@ the ReLU activation function has a single point where the derivative is not defined. As long as the linear map preceding the activation is not degenerate, then the non-smooth region has measure zero, and the change-of-variables technique still applies.

We cannot guarantee that the Jacobian has full rank through clever choices of neural architectures. However, we note that one requirement is that no hidden layer may have dimensionality below the $d$ dimensions of the latent space. This is a natural requirement for the generator anyway. In our model, we aim to maximize the entropy of the generator, which encourages the generator to create as diverse samples as possible. In practice this ensures that the Jacobian has full rank as a degenerate Jacobian implies a reduction of entropy. Note that this is not a theoretical guarantee against degenerate Jacobians during optimization, but in practice we have at no point experienced problems in this regard.

\subsection{Practical experimental settings}
For the toy and MNIST datasets, we use multi-layer perceptrons~(MLPs) networks, while for CIFAR-10 and ANIMEFACE datasets, we use a DCGAN network and Resnet architecture, respectively.
\begin{table}[htbp]
	\centering
	\caption{The network architecture trained for toy datasets}
		\begin{tabular}{lcc}
			\hline
			\hline
			Operation                       & \multicolumn{1}{l}{Input} & \multicolumn{1}{l}{Output} \\ \hline \hline
			\multicolumn{3}{c}{Energy}  
			 \\ \hline	\hline                       
			\multicolumn{1}{c}{Linear, PReLU} & 2                        & 100                         \\
			\multicolumn{1}{c}{Linear, PReLU} & 100                        & 100                        \\
			\multicolumn{1}{c}{Linear}          & 100                         & 1                          \\ \hline
			\hline
			\multicolumn{3}{c}{Generator}                                                              \\ \hline
			\hline
			\multicolumn{1}{c}{Linear, PReLU, BN }       & 2                         & 100                        \\
			\multicolumn{1}{c}{Linear, PReLU, BN }       & 100                         & 100                        \\ 
				\multicolumn{1}{c}{Linear}          & 100                         & 2                \\ \hline \hline
		\end{tabular}
	\end{table}

\begin{table}[htbp]
	\centering
	\caption{The network architecture trained for MNIST dataset}
		\begin{tabular}{lcc}
			\hline
			\hline
			\multicolumn{1}{c}{Operation}                       & \multicolumn{1}{c}{Input} & \multicolumn{1}{c}{Output} 
			\\ \hline \hline
			\multicolumn{3}{c}{Energy}  
			 \\ \hline	\hline                       
			\multicolumn{1}{c}{Linear, PReLU} & 2352                        & 2000                         \\
			\multicolumn{1}{c}{Linear, PReLU} & 2000                        & 1000                        \\
			\multicolumn{1}{c}{Linear, PReLU} & 1000                        & 500                        \\
			\multicolumn{1}{c}{Linear, PReLU} & 500                        & 250                        \\
			\multicolumn{1}{c}{Linear, PReLU} & 250                        & 250                        \\
			\multicolumn{1}{c}{Linear}          & 250                         & 1                          \\ \hline
			\hline
			\multicolumn{3}{c}{Generator}                                                              \\ \hline
			\hline
			\multicolumn{1}{c}{Linear, BN, PReLU }       & 128                         & 500                        \\
			\multicolumn{1}{c}{Linear, BN, PReLU }       & 500                         & 1000                        \\ 
			\multicolumn{1}{c}{Linear, BN, PReLU }       & 1000                         & 2000                        \\ 
			
				\multicolumn{1}{c}{Linear, Tanh}          & 2000                         & 2352                \\ \hline \hline
		\end{tabular}
	\end{table}
	
\begin{table}[htbp]
	\centering
	\caption{The network architecture for CIFAR-10 dataset}
	\label{network3}
	\begin{tabular}{ccccc}
		\hline \hline
		Operation                                     & \multicolumn{1}{l}{Kernel} & \multicolumn{1}{l}{Strides} & \multicolumn{1}{l}{Channels} & \multicolumn{1}{l}{Output size}\\ \hline \hline
		\multicolumn{4}{c}{Energy}                                                                      \\ \hline \hline
		Conv2D, LReLU                                  & 3                         & 1                & 64    & 32                           \\
		Conv2D, LReLU                               & 4                          & 2                           & 64      & 16                        \\
		Conv2D, LReLU                               & 3                         & 1                          & 128    & 16                          \\
		Conv2D, LReLU                               & 4                          & 2                           & 128   & 8                           \\
		Conv2D, LReLU                               & 3                          & 1                           & 256    & 8                          \\
		Conv2D, LReLU                                     & 4                          & 2                           & 256 & 4
		\\
		Conv2D, LReLU                       & 3                          & 1                           & 512   & 4                         \\
		Flatten   & & & & 8192
		\\
		Linear   & & & & 1
		\\ \hline \hline
		\multicolumn{4}{c}{Generator}                                                                                                                 \\ \hline \hline
		Linear & & & & 8192                             \\
		Reshape   &                           &                           & 512  & 4                             \\
		\multicolumn{1}{l}{ConvTranspose2D, BN, ReLU}   & 4                          & 2                           & 256 &  8                             \\
		\multicolumn{1}{l}{ConvTranspose2D, BN, ReLU}   & 4                          & 2                           & 128 & 16                               \\
		\multicolumn{1}{l}{ConvTranspose2D, BN, ReLU} & 4                          & 2        & 64      & 32
		\\
		Conv2D, Tanh & 3 & 1 & 3 & 32 
		\\ \hline \hline
	\end{tabular}
\end{table}

\begin{table}[htbp]
	\centering
	\caption{The energy network architecture for ANIMEFACE dataset}
	\label{energy network}
	\begin{tabular}{ccccc}
		\hline \hline
		Operation                                     & \multicolumn{1}{l}{Kernel} & \multicolumn{1}{l}{Strides} & \multicolumn{1}{l}{Channels} & \multicolumn{1}{l}{Output size}\\ \hline \hline
		(ResBlock0) \\
	  	\multicolumn{1}{l}{Left:  Conv2D, BN, LReLU}                                 & 3                         & 1                & 64    & 64                           \\
		Conv2D                                & 3                          & 1                           & 64      & 64                        \\
		AvgPool2D               & 2 & 2 & 64 & 32\\
		\multicolumn{1}{l}{Right: \quad 	AvgPool2D, BN}              & 2 & 2 & 3 & 32\\
		Conv2D                                & 1                         & 1                          & 64    & 32                          \\
		\multicolumn{1}{l}{Overall: \quad Add} & & &64 & 32\\ \hline
		(ResBlock1) \\
	  	\multicolumn{1}{l}{Left: BN, LReLU,Conv2D}                                 & 3                         & 1                & 128    & 32                           \\
		\multicolumn{1}{c}{BN, LReLU,Conv2D}                                 & 3                         & 1                & 128    & 32                           \\
		AvgPool2D               & 2 & 2 & 128 & 16\\
		\multicolumn{1}{l}{Right: \quad BN, Con2D}              & 1 & 1 & 128 & 32\\
		AvgPool2D                                & 2                         & 2                          & 128    & 16                          \\
		\multicolumn{1}{l}{Overall: \quad Add} & & &128 & 16\\ \hline
	    (ResBlock2) \\
	  	\multicolumn{1}{l}{Left: BN, LReLU,Conv2D}                                 & 3                         & 1                & 256    & 16                           \\
		\multicolumn{1}{c}{BN, LReLU,Conv2D}                                 & 3                         & 1                & 256    & 16                           \\
		AvgPool2D               & 2 & 2 & 256 & 8\\
		\multicolumn{1}{l}{Right: \quad BN, Con2D}              & 1 & 1 & 256 & 16\\
		AvgPool2D                                & 2                         & 2                          & 256    & 8                          \\
		\multicolumn{1}{l}{Overall: \quad Add} & & &256 & 8\\ \hline
		 (ResBlock3) \\
	  	\multicolumn{1}{l}{Left: BN, LReLU,Conv2D}                                 & 3                         & 1                & 512    & 8                           \\
		\multicolumn{1}{c}{BN, LReLU,Conv2D}                                 & 3                         & 1                & 512    & 8                          \\
		AvgPool2D               & 2 & 2 & 512 & 4\\
		\multicolumn{1}{l}{Right: \quad BN, Con2D}              & 1 & 1 & 512 & 8\\
		AvgPool2D                                & 2                         & 2                          & 512    & 4                         \\
		\multicolumn{1}{l}{Overall: \quad Add} & & &512 & 4\\ \hline
		 (ResBlock4) \\
	  	\multicolumn{1}{l}{Left: BN, LReLU,Conv2D}                                 & 3                         & 1                & 1024    & 4                           \\
		\multicolumn{1}{c}{BN, LReLU,Conv2D}                                 & 3                         & 1                & 1024    & 4                           \\
		\multicolumn{1}{l}{Right: \quad BN, Con2D}              & 1 & 1 & 1024 & 4\\
		\multicolumn{1}{l}{Overall: \quad Add} & & &1024 & 4\\ \hline
		LReLU   & & & 1024 & 4\\
		Sum      & & & & 1024\\
		Linear  & & & & 1
		\\ \hline \hline
		\end{tabular}
\end{table}
\begin{table}[htbp]
	\centering
	\caption{The generator network architecture for ANIMEFACE dataset}
	\label{generator network}
	\begin{tabular}{ccccc}
		\hline \hline
		Operation                                     & \multicolumn{1}{l}{Kernel} & \multicolumn{1}{l}{Strides} & \multicolumn{1}{l}{Channels} & \multicolumn{1}{l}{Output size}\\ \hline \hline
		Linear & & & & 16384                             \\
		Reshape   &                           &                           & 1024  & 4                             \\ \hline
		(ResBlock0) \\
	  	\multicolumn{1}{l}{Left: BN, ReLU, NN-Upsampling, Conv2D}                                 & 3        & 1                & 512    & 8                           \\
		BN, ReLU, Conv2D                                & 3                          & 1                           & 512      & 8                        \\
		\multicolumn{1}{l}{Right: \quad NN-Upsampling, Conv2D}              & 1 & 1 & 512 & 8\\
		\multicolumn{1}{l}{Overall: \quad  Add} & & &512 & 8 \\ \hline
		(ResBlock1) \\
	  	\multicolumn{1}{l}{Left: BN, ReLU, NN-Upsampling, Conv2D}                                 & 3        & 1                & 256    & 16                           \\
		BN, ReLU, Conv2D                                & 3                          & 1                           & 256      & 16                        \\
		\multicolumn{1}{l}{Right: \quad NN-Upsampling, Conv2D}              & 1 & 1 & 256 & 16 \\
		\multicolumn{1}{l}{Overall: \quad Add} & & &256 & 16 \\ \hline
		(ResBlock2) \\
	  	\multicolumn{1}{l}{Left: BN, ReLU, NN-Upsampling, Conv2D}                                 & 3        & 1                & 128    & 32                          \\
		BN, ReLU, Conv2D                                & 3                          & 1                           & 128      & 32                        \\
		\multicolumn{1}{l}{Right: \quad NN-Upsampling, Conv2D}              & 1 & 1 & 128 & 32 \\
		\multicolumn{1}{l}{Overall: \quad Add} & & &128 & 32 \\ \hline
		(ResBlock3) \\
	  	\multicolumn{1}{l}{Left: BN, ReLU, NN-Upsampling, Conv2D}     & 3        & 1     & 64    & 64                           \\
		BN, ReLU, Conv2D           & 3          & 1                           & 64      & 64                        \\
		\multicolumn{1}{l}{Right: \quad NN-Upsampling, Conv2D}              & 1 & 1 & 64 & 64 \\
		\multicolumn{1}{l}{Overall: \quad Add} & & & 64 & 64 \\ \hline
		BN, ReLU, Conv2D, Tanh   & 3 & 1 & 3 & 64
		\\ \hline \hline
	\end{tabular}
\end{table}

\section{Training Details}
In Table~\ref{parameter setting}, we specify the hyperparameters used when training our models for each dataset. We choose $p=2$ in Eq~(20) for our implementation. We normalize the data to be in [-1, 1] and do not use dequantization. During training we augment only using random horizontal flips. For cifar10 we used the official train-test split from PyTorch, and for animeface we used a 85:15 train-test split. For our upper bound, we set $\frac{M}{s_1^2}=\frac{0.001}{z_{dim}}$, where $s_1$ is the smallest singular value of Jacobian and $z_{dim}$ is the latent dimension. We observe this setting can get a satisfying generation for all datasets. If we replace the $z_{dim}$ with $\| v \|_2 ^2$ in high-dimensional data, where $v$ is a random vector sampled in Eq~19, it will further improve the generation. For out-of-distribution detection and capacity usage, we set $\frac{M}{s_1^2}=\frac{0.1}{z_{dim}}$, because if we increase this value, it will help the density estimation of the energy function. We also observe that the network's design affect performance. For example, removing batch normalization in the energy function can stabilize training on the Animeface dataset. The relationship between such design decisions in the context of EBMs should be explored in future work.
\begin{table}[htbp]
\centering
\caption{Selection of most important hyper-parameters and their
setting.}
\label{parameter setting}
\begin{tabular}{cccccc}
\hline
Datasets & Optimization & \multicolumn{1}{l}{Learning rate} & \multicolumn{1}{l}{Batch size} & \multicolumn{1}{l}{Iterations/Epochs} & \multicolumn{1}{l}{latent dim} \\ \hline
Toy & Adam(0.0,0.9) & 2e-4 & 200 & 150000 & 2 \\
MNIST & Adam(0.0,0.9) & 2e-4 & 64 & 60(epochs) & 128 \\
CIFAR-10 & Adam(0.0,0.999) & 5e-5 & 64 & 200000 & 128 \\
ANIMEFACE & Adam(0.0,0.999) & 5e-5 & 64 & 100000 & 128 \\ \hline
\end{tabular}
\end{table}


\end{document}